\documentclass{article} 
\usepackage{iclr2026_conference,times}

\usepackage{amsmath,amsfonts,bm}

\def\eqref#1{equation~\ref{#1}}

\def\1{\bm{1}}

\DeclareMathAlphabet{\mathsfit}{\encodingdefault}{\sfdefault}{m}{sl}
\SetMathAlphabet{\mathsfit}{bold}{\encodingdefault}{\sfdefault}{bx}{n}

\usepackage{hyperref}
\usepackage{url}

\usepackage{algorithm}
\usepackage{algorithmic}

\usepackage[table]{xcolor}
\usepackage{microtype}
\usepackage{graphicx}
\usepackage{subfigure}
\usepackage{booktabs} 

\usepackage{enumitem}
\usepackage{array}
\usepackage{amsmath}
\usepackage{siunitx} 
\usepackage{multirow}

\usepackage{bbding}
\usepackage{pifont}
\usepackage{wasysym}
\usepackage{amssymb}

\usepackage{xcolor}
\usepackage{colortbl}
\usepackage{rotating}
\usepackage{tabularx}
\usepackage{makecell}
\usepackage{placeins}  

\usepackage{svg}

\usepackage{pdflscape} 

\usepackage{wrapfig}

\usepackage[capitalize]{cleveref}

\usepackage{newfloat}
\usepackage{listings}

\title{Exploring Real-Time Super-Resolution: \\Benchmarking and Fine-Tuning
for Streaming Content}

\author{Evgeney Bogatyrev\textsuperscript{1,2,3}, Khaled Abud\textsuperscript{1,2,3}, Ivan Molodetskikh\textsuperscript{1,2}, Nikita Alutis\textsuperscript{1,2}, \\ \textbf{Dmitriy Vatolin}\textsuperscript{\textbf{1,2,3}} \\
\textsuperscript{1}Lomonosov Moscow State University, 119991, Moscow, Russia\\
\textsuperscript{2}AI Center, Lomonosov Moscow State University\\
\textsuperscript{3}MSU Institute for Artificial Intelligence, Lomonosov Moscow State University\\
\texttt{\
evgeney.bogatyrev@graphics.cs.msu.ru} \\
\texttt{\
khaled.abud@graphics.cs.msu.ru} \\
\texttt{\
ivan.molodetskikh@graphics.cs.msu.ru} \\
\texttt{\
nikita.alutis@graphics.cs.msu.ru} \\
\texttt{\
dmitriy@graphics.cs.msu.ru} \\
}

\iclrfinalcopy 
\begin{document}

\maketitle

\begin{abstract}
Recent advancements in real-time super-resolution have enabled higher-quality video streaming, yet existing methods struggle with the unique challenges of compressed video content. Commonly used datasets do not accurately reflect the characteristics of streaming media, limiting the relevance of current benchmarks. To address this gap, we introduce a comprehensive dataset - \textbf{StreamSR} - sourced from YouTube, covering a wide range of video genres and resolutions representative of real-world streaming scenarios. We benchmark 11 state-of-the-art real-time super-resolution models to evaluate their performance for the streaming use-case.

Furthermore, we propose \textbf{EfRLFN}, an efficient real-time model that integrates Efficient Channel Attention and a hyperbolic tangent activation function - a novel design choice in the context of real-time super-resolution. We extensively optimized the architecture to maximize efficiency and designed a composite loss function that improves training convergence. EfRLFN combines the strengths of existing architectures while improving both visual quality and runtime performance.

Finally, we show that fine-tuning other models on our dataset results in significant performance gains that generalize well across various standard benchmarks. We made the dataset, the code, and the benchmark available at \url{https://github.com/EvgeneyBogatyrev/EfRLFN}.
\end{abstract}

\section{Introduction}
\label{sec:intro}
In recent years, the demand for high-quality video streaming has surged, driven by the increasing adoption of platforms such as YouTube, Twitch, and Netflix~\citep{grandresearch}. To balance user experience with bandwidth constraints, streaming services often compress video content, which may introduce artifacts such as blockiness, blurring, and loss of fine details. While real-time super-resolution (SR) methods have emerged as a solution to enhance video quality, existing techniques struggle to address the unique challenges posed by heavily compressed videos in real-world scenarios~\citep{bogatyrev2023srcodec}.

Further investigations suggest that there are a limited number of SR methodologies capable of effectively upscaling compressed videos~\citep{li2021comisr}. Additionally, there is a notable scarcity of models designed for real-time video upscaling. The NTIRE challenges~\citep{Conde_2023_CVPR, ren2024ninth} have played a pivotal role in advancing the development of real-time SR models, fostering significant progress in this domain. However, it is worth mentioning that many of these SR models have been primarily trained on datasets such as DIV2K~\citep{Agustsson_2017_CVPR_Workshops} or Vimeo90K~\citep{xue2019video}, which do not adequately represent compressed video scenarios.

\begin{figure}[h!]
\centering
\includegraphics[width=0.99\columnwidth]{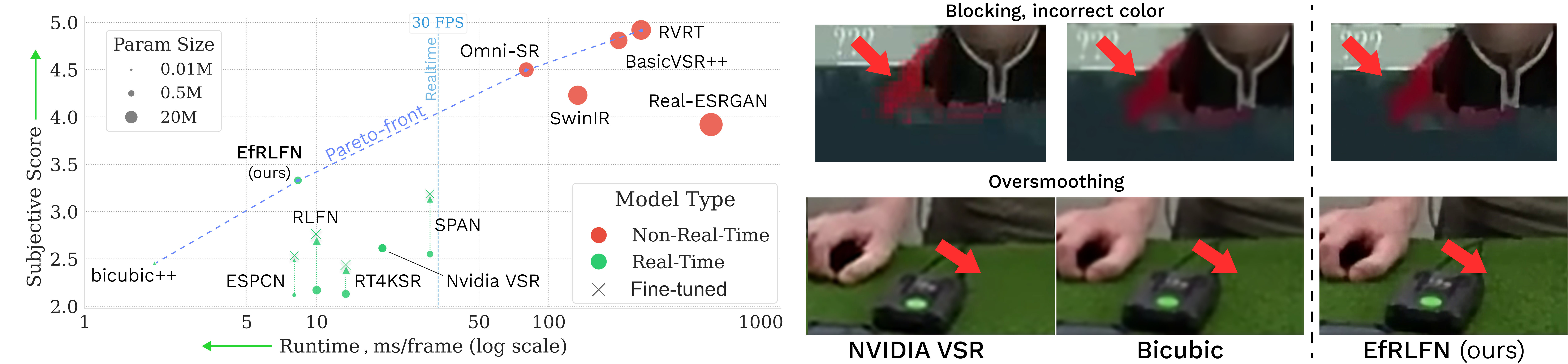}
\caption{\textbf{Left}: Trade-off between user preference score and runtime speed for various $2\times$ super-resolution models. Blue line represents Pareto-optimal front. Models achieving real-time performance\protect\footnotemark  are shown in green, while slower models are red. ``$\times$''  represent models fine-tuned on StreamSR dataset. \textbf{Right}: Examples of NVIDIA VSR artifacts compared to the proposed EfRLFN model and bicubic interpolation.}
\label{fig:scatter}
\end{figure}

NVIDIA’s Video Super-Resolution (NVIDIA VSR), a learning-based SR method integrated into GPU driver, demonstrates the growing focus on real-time SR for streaming applications~\citep{nvidia-vsr}. However, despite its efficiency, NVIDIA VSR often fails to restore fine textures and introduces over-smoothing, limiting its effectiveness, as illustrated in Figure \ref{fig:scatter}. Similarly, other state-of-the-art SR methods, such as SPAN~\citep{wan2024swift} and RLFN~\citep{kong2022residual}, demonstrate promising results on standard datasets but are not optimized for the complexities of streaming content~\citep{bogatyrev2023srcodec}. 

To address these challenges, we present \textbf{StreamSR}: a comprehensive dataset of 5,200 YouTube videos (and more than 10M individual frames) covering diverse genres and resolutions, offering a robust benchmark for evaluating SR methods under streaming conditions. Additionally, we propose \textbf{Efficient Residual Lightweight Feature Network (EfRLFN)}. This model is specifically optimized for real-time applications, combining techniques adopted from prior work in real-time SR and related fields. EfRLFN achieves significant improvements in terms of both objective and subjective video quality metrics while maintaining computational efficiency. 

Despite the promising results of video-based models on compressed video tasks, their architectures are usually too complex to meet the FPS requirements for real-time deployment. For this reason, the proposed EfRLFN model was intentionally developed as an image SR model, with the goal of combining effective methodologies and optimizing existing architectures to achieve superior performance.

Our \textbf{main contributions} are as follows:
\begin{enumerate}

\item \textbf{StreamSR – Streaming Super-Resolution Dataset}. We collect and analyze a multiscale dataset derived from YouTube, comprising 5,200 carefully filtered user-generated videos of diverse topics and content types, and conduct a comparative analysis with existing datasets.

\item \textbf{EfRLFN – Efficient Real-Time Super-Resolution Method}. Building upon a popular RLFN architecture, we thoroughly optimize it with a new block design, Efficient Channel Attention and enhanced training procedure. Our evaluations show that EfRLFN establishes a new SoTA in real-time SR, outperforming other methods in quality-complexity tradeoff, supported by extensive ablation experiments.

\item \textbf{Systematic benchmarking of real-time SR models}. We evaluate 11 real-time super-resolution methods using our dataset, assessing their performance both with and without fine-tuning on the training set. We employ 7 different objective image quality metrics and conduct a large-scale user preference study with more than 3,800 participants. User preference dataset will be also published alongside the SR benchmark.



\end{enumerate}

\footnotetext{$\ge$30 FPS on a Nvidia RTX 2080 GPU, which approximates a potential user setup}



\section{Related Work}
\label{sec:related_work}

\subsection{Super-Resolution Methods}

Recent advances in deep learning have significantly improved super-resolution (SR) techniques, with particular progress in real-time applications. Current approaches achieve varying balances between reconstruction quality and computational efficiency through innovative architectural designs.

Among real-time capable models, \textbf{Bicubic++}~\citep{bilecen2023bicubic++} combines fast reversible feature extraction with global structured pruning of convolutional layers, achieving processing speeds comparable to traditional bicubic interpolation. \textbf{AsConvSR}~\citep{guo2023asconvsr} introduces dynamic assembled convolutions that adapt their kernels based on input characteristics, optimized particularly for mobile and edge device deployment. For high-resolution upscaling, \textbf{RT4KSR}~\citep{zamfir2023towards} employs pixel-unshuffling and structural re-parameterization of NAFNet blocks to efficiently handle 720p/1080p to 4K conversion. The \textbf{SPAN} architecture~\citep{wan2024swift} features a parameter-free attention mechanism with symmetric activations to amplify important features and suppress redundant ones, reducing computational overhead by 40\% compared to conventional attention modules. Similarly, \textbf{RLFN}~\citep{kong2022residual} implements a streamlined three-layer convolutional structure within a residual learning framework, achieving 1080p to 4K conversion in under 15ms while maintaining strong feature aggregation. \textbf{TMP}~\citep{zhang2024tmp} enhances online video super-resolution by minimizing computational redundancy in motion estimation.

In addition to lightweight real-time methods, transformer-based models like \textbf{SwinIR}~\citep{liang2021swinir} and \textbf{HiT-SR}~\citep{zhang2025hit} achieve superior quality through self-attention mechanisms and hierarchical multi-scale processing. There are other approaches to SR task, such as \textbf{COMISR}~\citep{li2021comisr}, which specialize in compressed video upscaling. \textbf{Real-ESRGAN}~\citep{wang2021real} is a popular GAN-based method that employs high-order degradation modeling to simulate real-world degradation. Although these methods demonstrate superior results compared to real-time SR models, they are typically computationally intensive, making them unsuitable for real-time applications.

A growing line of research specifically targets compression-aware video super-resolution, where models are designed to handle distortions introduced by modern video codecs. Representative works include \textbf{CAVSR}~\citep{wang2023compression} and \textbf{TAVSR}~\citep{tavsr}, which leverage codec-related features to restore severely degraded compressed sequences. Recent approaches such as \textbf{FTVSR}~\citep{ftvsr} and \textbf{FTVSR++}~\citep{ftvsr++} utilize transformer architectures that perform self-attention across joint space–time–frequency domains. Other works~\citep{diffcomp} incorporate diffusion-based denoising and distortion-control modules to reduce the impact of compression artifacts during generation. These approaches typically achieve strong reconstruction quality on low-bitrate inputs but rely on computationally heavy architectures, making them unsuitable for real-time deployment.

While all these models have demonstrated varying degrees of success in enhancing image quality through super-resolution techniques, challenges remain in effectively addressing the issues posed by compressed video content. In this paper, we build upon this foundational works aiming to evaluate and incorporate best practices to create the most suited SR model for real-time use case.

\subsection{Datasets and Real-time SR Challenges}

\begin{figure*}[t!]
  \centering
  \includegraphics[width=0.95\linewidth]{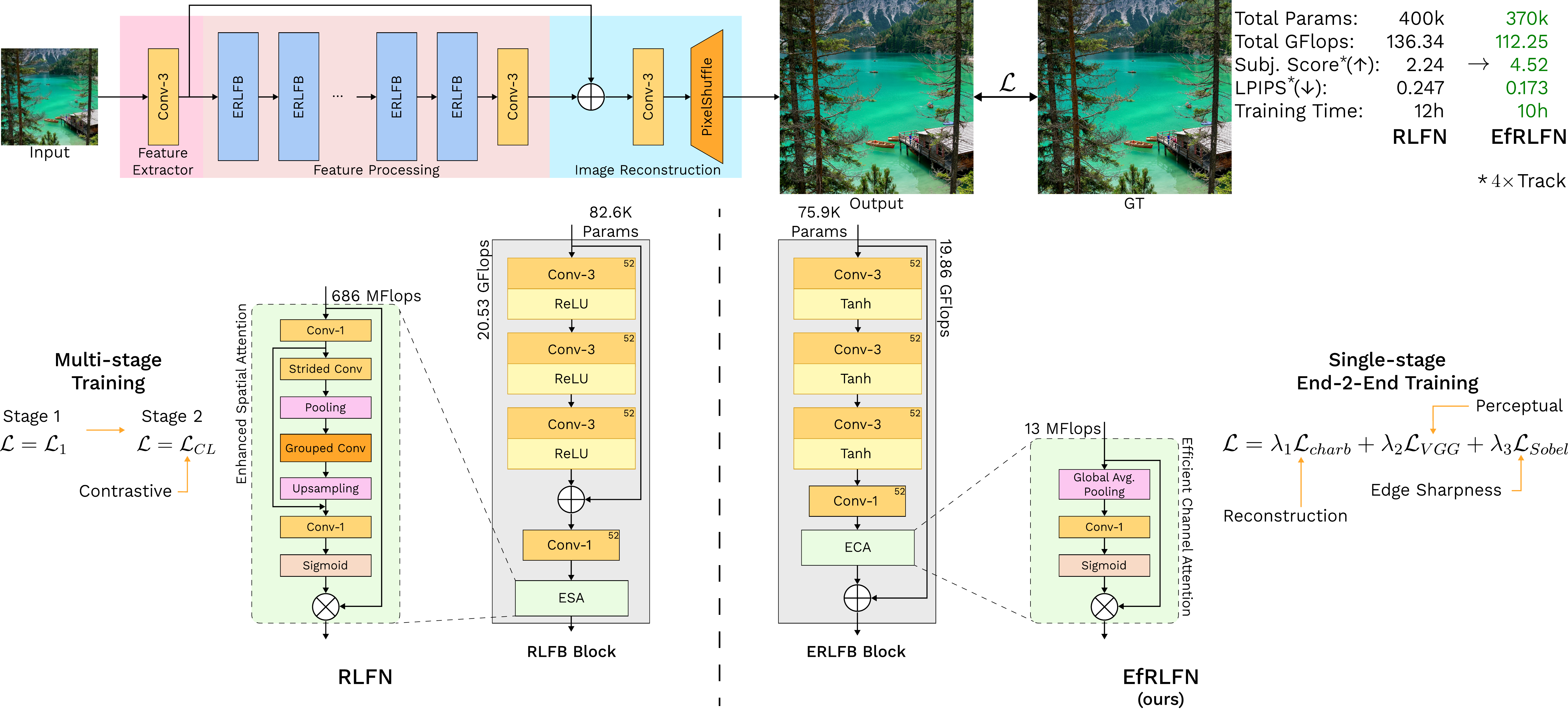}
\caption{Visual summary of the proposed EfRLFN model and the comparison with the original RLFN architecture.}
\label{fig:model-arch}
\end{figure*}

Datasets play a crucial role in training SR models. While \textbf{DIV2K}~\citep{Agustsson_2017_CVPR_Workshops} and \textbf{Vimeo90K}~\citep{xue2019video} provide clean HR-LR pairs, they lack streaming compression artifacts. The \textbf{YouTube-8M} dataset~\citep{abu2016youtube} offers diverse content but not multi-scale real-time scenarios. Additionally, several benchmark datasets are widely used for evaluating SR performance, including Set14~\citep{zeyde2012single}, BSD100~\citep{bsd100}, and Urban100~\citep{huang2015single}, which contain natural and structured images at fixed scales. For video super-resolution tasks, REDS~\citep{nah2019ntire} provides dynamic scenes with motion blur and compression artifacts, making it suitable for training and testing temporal SR models. 

Recent SR advancements focus on real-time efficiency. The \textbf{NTIRE 2023 Real-Time SR Challenge}~\citep{Conde_2023_CVPR} promoted lightweight architectures using pruning, quantization, and knowledge distillation for live applications. Its successor, \textbf{NTIRE 2024 Efficient SR challenge}~\citep{ren2024ninth}, emphasized resource efficiency with adaptive transformers and multi-task learning combining SR with denoising and deblocking. The latest \textbf{NTIRE 2025 challenge}~\citep{ren2025tenth} further advanced efficiency by balancing runtime, FLOPs, and parameters, introducing innovations like state-space models and ConvLora-based distillation.

While these challenges rank real-time SR methods, their exclusion of compressed video limits their applicability to streaming content. With this idea in mind, we created a novel dataset which includes diverse videos with compression at several resolutions.

\section{Proposed Model}

\subsection{Network Architecture}

Although video SR models excel in standard SR tasks, their high complexity hinders real-time processing and batch training. Thus, we focus on image SR for our approach. The proposed \textbf{Efficient Residual Local Feature Network (EfRLFN)} enhances the RLFN~\citep{kong2022residual} architecture through targeted modifications to improve efficiency and reconstruction quality. Figure \ref{fig:model-arch} demonstrates the overall architecture of the model and its core modifications over RLFN. EfRLFN consists of a feature extraction block, main feature processing module, and an output reconstruction block. Due to their simplicity and parameter efficiency, we retain the feature extractor and reconstruction modules from the original RLFN model. The key innovations of our model are focused on ERLFB block design and the training procedure. 

\textbf{Efficient Residual Local Feature Blocks (ERLFB)} replace the original RLFB blocks in RLFN. The primary change involves adopting \textbf{tanh activation} in the refinement modules instead of ReLU. This modification is motivated by recent findings \citep{wan2024swift} demonstrating that odd symmetric activation functions, such as $Sigmoid(x) - 0.5$ or tanh, preserve both the magnitude and sign of features. Unlike ReLU, which discards negative activations, tanh ensures richer gradient flow and prevents directional information loss in attention maps, leading to more accurate feature refinement. Empirical studies demonstrate that tanh-based activation can enhance gradient propagation in deep networks~\citep{clevert} and improve performance in dense prediction tasks, including image-to-image translation~\citep{efros, esrgan}.

Another important modification lies in the attention mechanism used for global information aggregation. \textbf{Efficient Channel Attention (ECA)} substitutes the Enhanced Spatial Attention (ESA) mechanism used in RLFN. While ESA employs multi-layer convolutional groups for spatial attention, ECA leverages a lightweight channel attention mechanism based on global average pooling and a $1\times1$ convolution. This change significantly reduces computational overhead while maintaining competitive performance, as validated in our ablation studies.

Additionaly, the \textbf{reconstruction path} of the ERLFB is streamlined by removing redundant skip connections and simplifying the feature smoothing step to a single $3\times3$ convolution. This optimization reduces computational fragmentation, improving inference speed without compromising reconstruction quality.

\subsection{Loss Function}
\begin{figure*}[t!]
  \centering
  \includegraphics[width=1.0\linewidth]{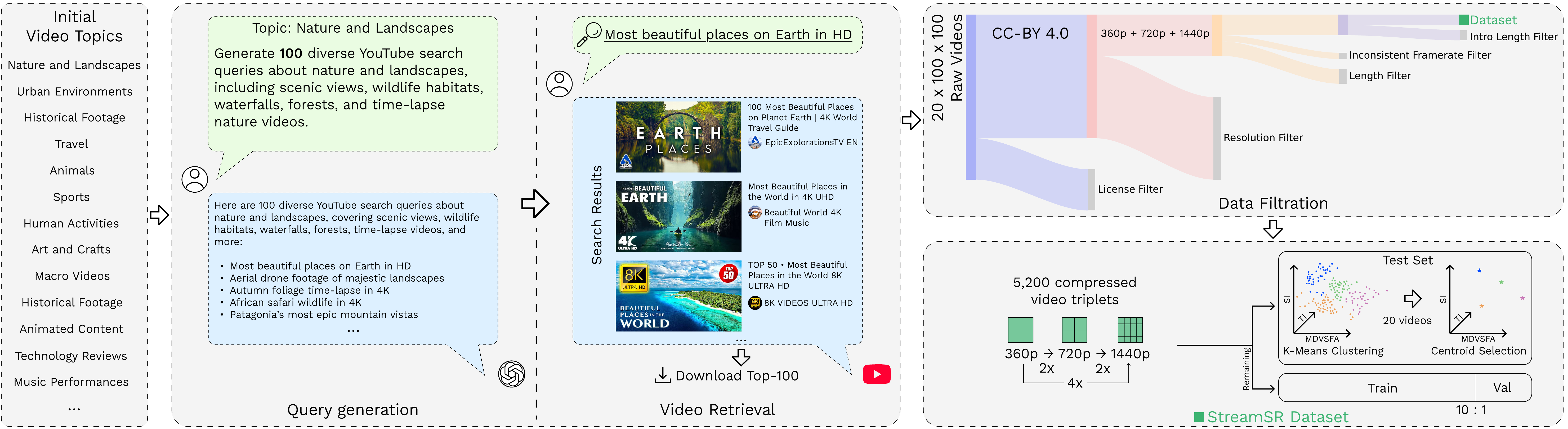}
\caption{The process of video collection for our StreamSR dataset. The resulting set is split into train, test, and validation parts. \textbf{Zoom in for better clarity.}}
\label{fig:dataset-scheme}
\end{figure*}

\begin{table*}[t]
\centering
\caption{Comparison of UGC Video Datasets. Quality range shows 5th-95th percentile scores of MDTVSFA metric. Duration shows the typical duration of a video clip. \textit{Image} means that dataset contains only single images.}
\resizebox{0.95\textwidth}{!}{
\begin{tabular}{@{}lcccccc>{\raggedright\arraybackslash}p{2cm}@{}}
\toprule
\textbf{Dataset} & \textbf{Year} & \textbf{Size} & \textbf{Resolutions} & \textbf{LR?} & \textbf{Quality Range} & \textbf{Use-Case} & \textbf{Duration} \\
\midrule
YouTube-8M~\citep{abu2016youtube} & 2016 & 8M videos & 240p--1080p & \textcolor{red}{\texttimes} & 0.45--0.58 & Classification & 3--5 min \\
Kinetics-400~\citep{kay2017kinetics} & 2017 & 240k clips & 1080p & \textcolor{red}{\texttimes} & 0.48--0.59 & Action Recognition & 8--12 sec \\
Kinetics-700~\citep{carreira2019short} & 2019 & 650k clips & 1080p & \textcolor{red}{\texttimes} & 0.47--0.58 & Action Recognition & 8--12 sec \\
YouTube-VOS~\citep{xu2018youtube} & 2018 & 4.5k videos & 360p--1080p & \textcolor{red}{\texttimes} & 0.46--0.57 & Object Segmentation & 15--25 sec \\
YouTube-BB~\citep{real2017youtube} & 2017 & 380k videos & 1080p & \textcolor{red}{\texttimes} & 0.50--0.60 & Object Tracking & 8--12 sec \\
YT-Temporal 180M~\citep{zellersluhessel2021merlot} & 2020 & 180M clips & 144p--1080p & \textcolor{red}{\texttimes} & 0.42--0.56 & Language Pretraining & 3--7 sec \\
DIV2K~\citep{Agustsson_2017_CVPR_Workshops} & 2017 & 1,000 images & 2K & \textcolor{green!70!black}{\checkmark} & 0.55--0.65 & Image SR & \textit{Image} \\
REDS~\citep{nah2019ntire} & 2019 & 300 clips &  360p--720p & \textcolor{green!70!black}{\checkmark} & 0.50--0.60 & Video SR & 2–5 sec \\
Vimeo-90K~\citep{xue2019video} & 2019 & 90K clips & 448p &  \textcolor{red}{\texttimes} & 0.48--0.58 & Video SR & ~1 sec \\
UDM10~\citep{yi2019progressive} & 2019 & 10K pairs & 540p–4K  & \textcolor{green!70!black}{\checkmark} & 0.45--0.58 & Image SR & \textit{Image} \\
RealSR~\citep{cai2019toward} & 2019 & 595 pairs & 4K & \textcolor{green!70!black}{\checkmark} & 0.43-0.59 & Real-world SR & \textit{Image} \\
VideoLQ~\citep{chan2022basicvsr++} & 2021 & 1,000 clips & 270p–1080p & \textcolor{green!70!black}{\checkmark} & 0.43--0.54  & Video SR & 5–10 sec \\
\rowcolor{blue!5}
\textbf{StreamSR (Proposed)} & 2025 & 5.2k videos & 360p--1440p & \textcolor{green!70!black}{\checkmark} & \textbf{0.41--0.61} & Super-Resolution & 25--30 sec \\
\bottomrule
\end{tabular}
}
\label{tab:youtube_sr_datasets}
\end{table*}

While RLFN's contrastive loss effectively aligns intermediate features through positive and negative sample pairs, it introduces two practical challenges: (1) sensitivity to the choice of feature extractor layers (shallow layers are required for PSNR-oriented tasks \citep{kong2022residual}), and (2) increased computational overhead due to pairwise feature comparisons. EfRLFN introduces a composite loss function designed to address the limitations of RLFN's contrastive loss while emphasizing structural fidelity and edge preservation. The loss consists of three key components:

\begin{align}
\mathcal{L} = \underbrace{\lambda_{Charb} \mathcal{L}_{Charb}}_{\text{Reconstruction}} + \underbrace{\lambda_{VGG} \mathcal{L}_{VGG}}_{\text{Perception}} + \underbrace{\lambda_{Sobel} \mathcal{L}_{Sobel}}_{\text{Edge Sharpness}}
\end{align}

\textbf{Charbonnier Loss} enforces pixel-level accuracy while providing robustness to outliers, addressing a limitation of the L1/L2 losses used in RLFN's early training stages. Its square-root formulation ensures stable gradients even for large residuals, making it particularly suitable for super-resolution tasks.

\begin{align}
\mathcal{L}_{Charb} = \sqrt{\| \mathbf{I}_{HR} - \mathbf{I}_{SR} \|^2 + \epsilon^2}, \quad \epsilon > 0
\end{align}
where $I_{\text{HR}}$ denotes the reference high-resolution image and $I_{\text{SR}}$ denotes the output of the SR module.  
    
\textbf{VGG Perceptual Loss} leverages deep features from a pre-trained VGG-19 network~\citep{VGG} to guide the model toward perceptually realistic outputs.
\begin{align}
\mathcal{L}_{VGG} = \| \phi_{\text{VGG}}(\mathbf{I}_{HR}) - \phi_{\text{VGG}}(\mathbf{I}_{SR}) \|_1,
\end{align}
where \(\phi_{\text{VGG}}(\cdot)\) denotes ReLU activations from the \textit{conv5\_4} layer. While RLFN employs a contrastive loss to align intermediate features, our VGG loss provides analogous perceptual supervision without requiring complex feature pairing or negative sampling. This simplification reduces training complexity while maintaining high visual quality.

\textbf{Sobel Edge Loss} explicitly optimizes edge sharpness by penalizing discrepancies in gradient maps between the super-resolved and ground truth images. 

\begin{align}
\mathcal{L}_{Sobel} = \| S(\mathbf{I}_{HR}) - S(\mathbf{I}_{SR}) \|_2^2,
\end{align}
where \(S(\cdot)\) applies the Sobel operator to extract horizontal and vertical edges.
\begin{figure*}[h!]
\includegraphics[width=1.0\linewidth]{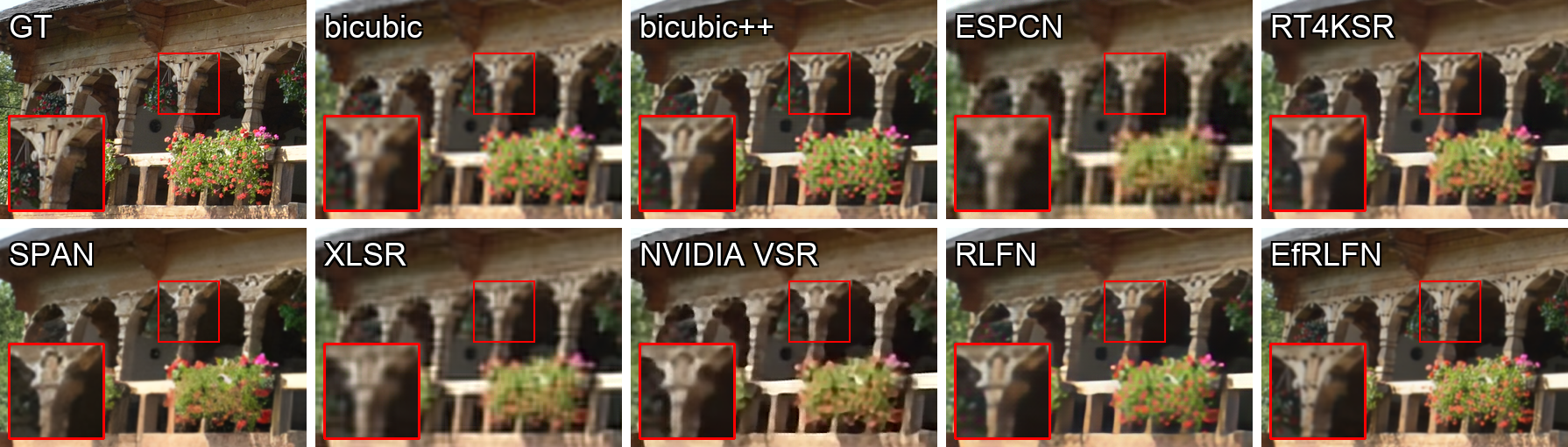}
\caption{The comparison between EfRLFN and several 4$\times$ real-time SR models.}
\label{fig:real-time-sr-compare-main}
\end{figure*}

This component addresses a critical gap in RLFN's approach, which relies on implicit edge preservation through contrastive learning on shallow features. By directly targeting edge quality, our loss ensures crisper and more confident reconstructions, particularly for high-frequency details. User studies confirm that such edge-enhanced outputs align with human perceptual preferences, achieving significantly higher subjective ratings~\citep{zhang2018unreasonable, bogatyrev2023srcodec}. 
 

Overall, EfRLFN outperforms RLFN with 15\% faster inference via hardware-friendly ECA and tanh activations, faster and more stable training through a simplified single-stage training with a carefully designed composite loss, and better quality as measured with objective and subjective metrics. We further discuss the impact of individual design choices in the Ablation Study.

\section{Real-time SR Benchmark}
\label{sec:dataset}

In this section, we detail the methodology employed for video collection process for the StreamSR dataset (summarized in Figure \ref{fig:dataset-scheme}), their subsequent categorization, and overall evaluation protocol and benchmark setup. 

\subsection{Dataset Collection}
\label{sec:dataset-collection-llm}

\textbf{Video Retrieval}\:\: We employed a Large Language Model (GPT-4o)~\citep{achiam2023gpt} to generate search queries for collecting videos. Starting with 20 distinct YouTube categories (e.g., travel, education, gaming), the LLM produced 100 diverse YouTube search queries per category. We then downloaded the top 100 videos for each query.  The list of initial topics, LLM prompts and examples of queries are provided in the Section \ref{sec:LLM}).

\noindent\textbf{Filtering}\:\: Only Creative Commons (\textit{CC BY 4.0}) videos were included, with available resolutions of 360p, 720p, and 1440p. Videos had to maintain a consistent frame rate, verified automatically, to enable precise frame alignment. Resolutions were selected to support two SR benchmark tracks: 2$\times$ (720p→1440p) and 4$\times$ (360p→1440p). For each resolution, the highest-bitrate stream was chosen to ensure optimal quality of the target high-res stream. We extracted the first 30 seconds of each video ($\le$2,000 frames). 
Videos with long static intros were discarded with a scene-difference filter analyzing the $1^{st}$, $100^{th}$ and $150^{th}$ frames. 
This yielded a collection of 5,200 videos, which we named \textbf{StreamSR} dataset.

\noindent\textbf{Categorization}\:\: To construct a diverse and representative test set, we clustered videos using SI, TI~\citep{wang2019youtube}, bitrate, MDTVSFA~\citep{li2021unified}, and SigLIP~\citep{zhai2023sigmoid} text embeddings of the search queries. SigLIP embeddings were projected into 3-dimentional space with PCA. From each of the 20 resulting clusters, the video closest to the centroid was selected. The rest were split into training and validation sets in a 10:1 ratio.


Table~\ref{tab:youtube_sr_datasets} compares our StreamSR dataset with existing UGC datasets. Unlike classification (YouTube-8M, Kinetics) or segmentation datasets (YouTube-VOS), StreamSR is specifically designed for super-resolution, providing aligned 360p-1440p videos with 2×/4× upscaling tracks. With longer 25-30s clips, LR-HR pairs with natural compression degradations, and broader quality range, it better represents real-world streaming scenarios than previous SR datasets (REDS, Vimeo-90K).

\subsection{Benchmark Setup}
\begin{table*}
\caption{Comparison of real-time SR methods with 95\% confidence intervals on $2\times$ super-resolution track. Best results are in \textbf{bold}, second best are \underline{underlined}. ``$T$'' indicates fine-tuning on StreamSR. ``Subj.'' represents the Bradley-Terry subjective scores. Extended results are in Section \ref{app:a10-full-results}.}
\centering
\tiny
\begin{tabular}{lcccccccr}
\toprule
 & \multicolumn{4}{c}{\textbf{Our Benchmark}} & \multicolumn{3}{c}{\textbf{Standard Benchmarks}} \\
\cmidrule(lr){2-5} \cmidrule(lr){6-8} 
\textbf{Method} & Subj.$\uparrow$ & PSNR$\uparrow$ & LPIPS$\downarrow$ & CLIP-IQA$\uparrow$ & BSD100 & Urban100 & DIV2K & FPS$\uparrow$ \\
 &  &  &  &  & SSIM$\uparrow$/LPIPS$\downarrow$ & SSIM$\uparrow$/LPIPS$\downarrow$ & SSIM$\uparrow$/LPIPS$\downarrow$ \\
\midrule
AsConvSR$^{T}$ & 1.93±0.11 & 35.25±0.15 & 0.214±0.008 & 0.48±0.02 & 0.832 / 0.270 & 0.824 / 0.208 & 0.883 / 0.206 & 213±8 \\
RT4KSR & 2.40±0.12 & 36.45±0.16 & 0.213±0.009 & 0.49±0.02 & 0.805 / 0.296 & 0.769 / 0.244 & 0.860 / 0.221 & 102±5 \\
RT4KSR$^{T}$ & 2.43±0.11 & 37.55±0.14 & 0.070±0.003 & 0.53±0.02 & 0.812 / 0.254 & 0.766 / 0.223 & 0.864 / 0.186 & 102±5 \\
bicubic & 2.44±0.14 & 30.32±0.21 & 0.076±0.005 & 0.44±0.03 & 0.752 / 0.386 & 0.711 / 0.324 & 0.820 / 0.292 & \cellcolor{blue!20}\textbf{1829±50} \\
ESPCN & 2.48±0.12 & 30.71±0.18 & 0.078±0.004 & 0.45±0.02 & 0.774 / 0.534 & 0.505 / 0.437 & 0.514 / 0.515 & 201±7 \\
ESPCN$^{T}$ & 2.49±0.11 & 35.76±0.15 & 0.072±0.003 & 0.51±0.02 & 0.814 / 0.236 & 0.840 / 0.150 & 0.862 / 0.180 & 201±7 \\ 
Bicubic++$^{T}$ & 2.44±0.13 & 36.79±0.15 & 0.087±0.004 & 0.52±0.02 & 0.768 / 0.360 & 0.725 / 0.298 & 0.832 / 0.270 & \underline{1629±45} \\
NVIDIA VSR & 2.57±0.14 & 37.40±0.13 & 0.082±0.003 & 0.56±0.02 & 0.788 / 0.243 & 0.786 / 0.152 & 0.858 / 0.180 & 52±3 \\
XLSR$^{T}$ & 2.56±0.12 & 37.25±0.14 & 0.230±0.010 & 0.47±0.02 & 0.817 / 0.244 & 0.828 / 0.158 & 0.864 / 0.183 & 429±15 \\
SMFANet$^{T}$ & 2.37±0.14 & 37.14±0.15 & 0.158±0.007 & 0.51±0.02 & 0.803 / 0.239 & 0.798 / 0.153 & 0.865 / 0.177 & 327±12 \\
SAFMN$^{T}$ & 2.26±0.15 & 37.09±0.15 & 0.122±0.006 & 0.53±0.02 & 0.813 / 0.236 & 0.808 / 0.150 & 0.871 / 0.175 & 273±10 \\
RLFN & 2.17±0.15 & 37.03±0.15 & 0.086±0.004 & 0.54±0.02 & 0.805 / 0.238 & 0.803 / 0.147 & 0.876 / 0.175 & 225±8 \\
RLFN$^{T}$ & 2.69±0.13 & 37.63±0.12 & 0.072±0.003 & 0.58±0.02 & 0.834 / 0.239 & 0.845 / 0.153 & 0.881 / 0.178 & 225±8 \\
SPAN & 2.55±0.12 & 37.45±0.13 & 0.066±0.003 & 0.57±0.02 & 0.836 / \underline{0.222} & 0.841 / 0.139 & 0.887 / \underline{0.168} & 60±3  \\
SPAN$^{T}$ & \underline{3.13±0.15} & \underline{37.73±0.12} & \underline{0.063±0.003} & \underline{0.61±0.02} & \underline{0.837} / 0.239 & \underline{0.847} / \underline{0.118} & \underline{0.890} / 0.175 & 60±3 \\

\textbf{EfRLFN}$^{T}$ & \cellcolor{blue!20}\textbf{3.33±0.14} & \cellcolor{blue!20}\textbf{37.85±0.11} & \cellcolor{blue!20}\textbf{0.059±0.003} & \cellcolor{blue!20}\textbf{0.65±0.02} & \cellcolor{blue!20}\textbf{0.847} / \cellcolor{blue!20}\textbf{0.190} & \cellcolor{blue!20}\textbf{0.856} / \cellcolor{blue!20}\textbf{0.116} & \cellcolor{blue!20}\textbf{0.892} / \cellcolor{blue!20}\textbf{0.145} & 271±10 \\
\midrule
\multicolumn{9}{l}{\textit{Non-real-time SR models}} \\
Real-ESRGAN & 3.87±0.13 & 37.65±0.12 & 0.048±0.007 & 0.66±0.02 & 0.857 / 0.172 & 0.867 / 0.112 & 0.901 / 0.137 & 9±1 \\
BasicVSR++ & 4.87±0.13 & 38.05±0.13 & 0.037±0.007 & 0.70±0.02 & 0.860 / 0.158 & 0.868 / 0.110 & 0.907 / 0.131 & 15±1 \\
COMISR & 4.32±0.14 & 38.15±0.12 & 0.033±0.006 & 0.67±0.03 & 0.894 / 0.161 & 0.871 / 0.108 & 0.913 / 0.125 & 7±1 \\
\bottomrule
\end{tabular}
\label{tab:combined_2x_results}
\end{table*}

For benchmarking and fine-tuning, we selected a comprehensive list of SR models based on the year-to-year results of the NTIRE challenges, ranging from classical approaches like bicubic, bicubic++~\citep{bilecen2023bicubic++} and ESPCN~\citep{talab2019super} up to the recent SoTA methods like SMFANet~\citep{smfanet}, RLFN~\citep{kong2022residual} and SPAN~\citep{wan2024swift}. We had also included a proprietary NVIDIA VSR model to the comparison. Table \ref{tab:model_specs} in the Appendix summarizes the information about the tested models.

We selected a diverse set of objective metrics that assess different aspects of SR algorithms' performance. PSNR and SSIM~\citep{wang2004image} serve as fundamental measures of pixel-level fidelity and structural similarity. LPIPS~\citep{zhang2018unreasonable}, on the other hand, goes beyond traditional metrics, and employs deep features to better align with human perception. Aside from Full-Reference metrics, we also included No-Reference (NR) models in the comparison. They assess the quality of super-resolved images independently, without comparing them to the real high-resolution examples. Specifically, we employed 2 NR metrics: MUSIQ~\citep{ke2021musiq}, a powerful transformer-based model that excels in real-world content assessment, and CLIP-IQA~\citep{wang2023exploring}, a popular model that leverages  vision-language correspondence to capture semantic preservation in SR outputs. We evaluate the alignment of these metrics with human preferences in Section \ref{sec:metric_corrs}, proving their applicability to the super-resolution quality assessment task.

The subjective comparison still remains the best option for evaluating SR performance. Since user studies can often be expensive, we preliminary excluded models that demonstrated subpar performance based on objective metrics. To evaluate and rank the performance of various SR models from a subjective perspective, we conducted a pair-wise crowd-sourced comparison utilizing the Subjectify.us~\citep{subjectify} service. Comparison included 11 SR models (+ Ground Truth) and 660 video pairs in total.

A total of 3,822 unique individuals participated in our subjective evaluation. Consequently, the final subjective scores were derived from the 37,184 valid responses, applying the Bradley-Terry model~\citep{bradley1952rank}. More detailed information on this study is provided in Section \ref{app:a5-crowcourcing}.

\section{Experiments}
\label{sec::exp}

\paragraph{Fine-tuning}
\label{seq:sr-select}

To ensure a fair benchmarking of each model, we pre-trained all tested models on the training set of our dataset and evaluated the results of both pre-trained and non-pre-trained checkpoints. If a model did not have pre-trained weights for a specific upscale factor (2$\times$ or 4$\times$), we trained it from scratch and utilized only the trained version. Additional information on the training can be found in Section \ref{app:a3-experimental-setup}.

\paragraph{Validation on the StreamSR}

\begin{figure}[t!]
\centering
\includegraphics[width=1.0\columnwidth]{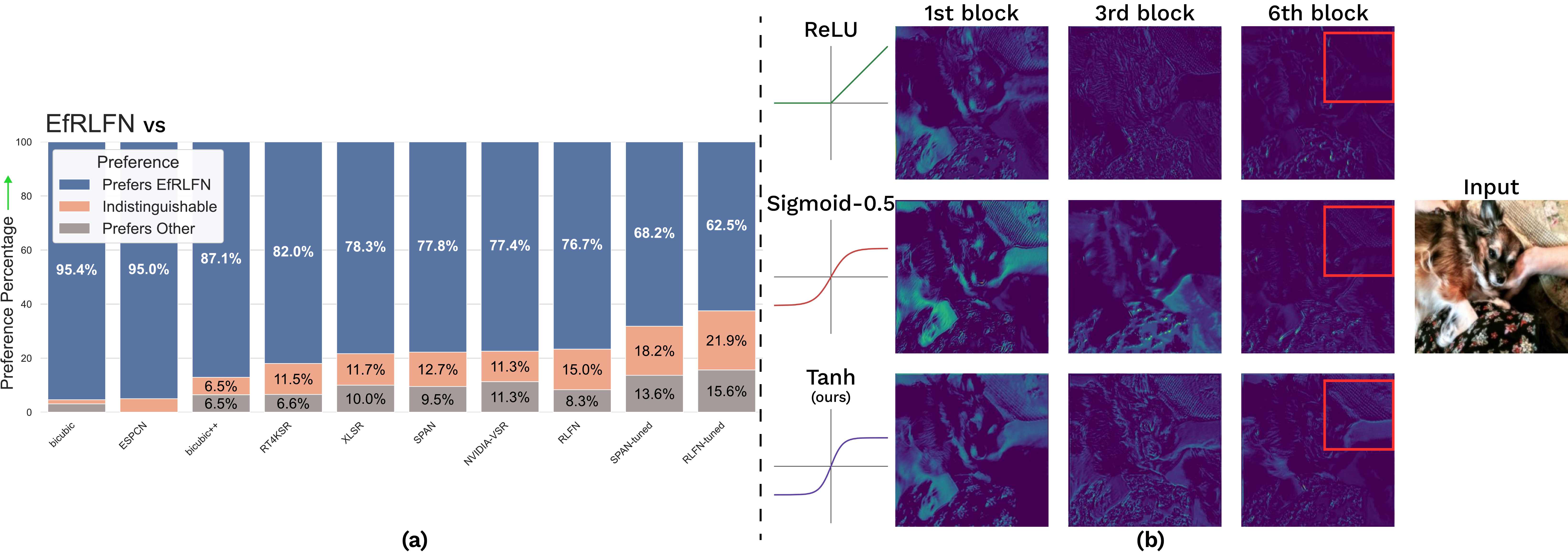}
\caption{(a) Pairwise preference evaluation of EfRLFN against other real-time super-resolution methods. (b) A comparison of the output feature maps from the first, third, and sixth ERLFB blocks. The features are taken from the output of the ECA block within each ERLFB. 
}
\label{fig:preference}
\end{figure}

Table~\ref{tab:combined_2x_results} (left) shows that the proposed model, EfRLFN, achieves superior performance on the test set of our dataset among the real-time models. The performance increase generalizes well across various objective metrics, and is further confirmed by the user study (Figure \ref{fig:preference}(a)). 
The results also indicate that SPAN, RLFN, and ESPCN models benefit significantly from fine-tuning on our dataset: there is a notable positive performance gap in both objective and subjective evaluations. After tuning, some of these models outperform the proprietary NVIDIA VSR model in user preferences (Figure \ref{fig:scatter}). We also include some non-real-time SR models for comparison.

Figure \ref{fig:preference}(a) demonstrates the results of pair-wise subjective comparison with other SR models: evidently, users largely prefer the outputs of EfRLFN over other methods. Specifically, EfRLFN is favored over NVIDIA VSR in 77.4\% of cases. The only competitive alternatives (SPAN-tuned and RLFN-tuned) are also trained on the StreamSR training set.

\paragraph{Validation on Other Datasets}

We validated EfRLFN and other real-time SR methods on five widely recognized datasets: Set14~\citep{zeyde2012single}, BSD100~\citep{bsd100}, Urban100~\citep{huang2015single}, REDS~\citep{nah2019ntire}, and the validation set of DIV2K~\citep{Agustsson_2017_CVPR_Workshops}. The average metric results are presented in Table \ref{tab:combined_2x_results} (right). EfRLFN consistently outperforms all competing approaches across all evaluated scenarios, achieving superior performance in both 2$\times$ and 4$\times$ resolution upscaling. Other quality assessment metrics, as well as the 4$\times$ dataset tracks are reported in Section \ref{app:a10-full-results}.

\begin{figure*}[h!]
\centering
\includegraphics[width=0.98\linewidth]{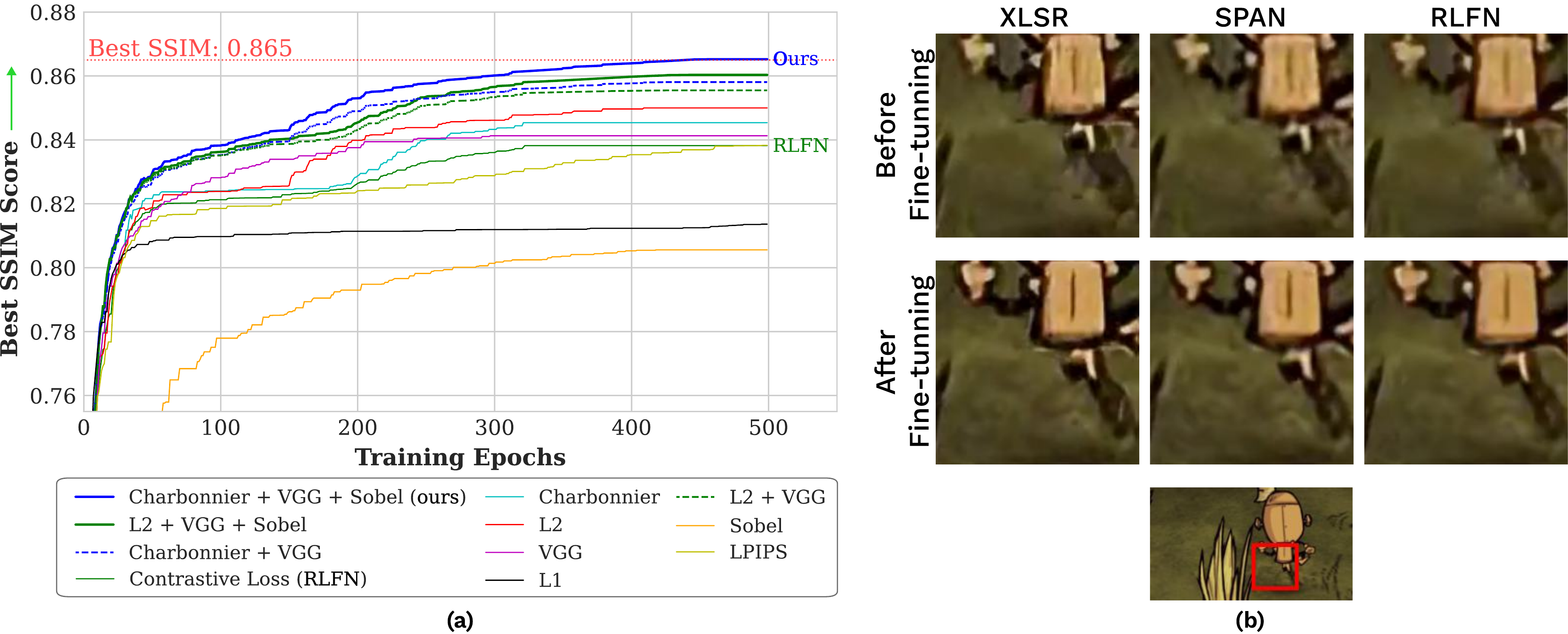}
\caption{(a) The impart of loss functions on the training of EfRLFN on $4\times$ track of the proposed dataset. ``Contrastive Loss'' represents the original training procedure of RLFN model. (b) The comparison between SR models before and after fine-tuning on our StreamSR dataset.}
\label{fig:losses-comparison}
\end{figure*}

Figures \ref{fig:real-time-sr-compare-main} and \ref{fig:losses-comparison}(b) provide a visual comparison between different models, further demonstrating the effectiveness of EfRLFN in real-time super-resolution. Visual comparisons clearly demonstrate EfRLFN's advantages: it produces sharper images with more accurate details, while competing models often struggle to reproduce clear boundaries and fine textures, resulting in noticeable artifacts or blurring in challenging regions.

Our experiments reveal that training on our proposed dataset not only benefits EfRLFN but also improves the performance of other competing models, as evidenced by the quantitative results in Table \ref{tab:combined_2x_results} (particularly noticeable in cases such as SPAN and RLFN).


\subsection{Ablation Study}
\label{sec:ablation}

\textbf{Loss Analysis}\:\:
Figure~\ref{fig:losses-comparison}(a) illustrates the impact of different loss functions on the SSIM performance of EfRLFN during training on the 4× track of our dataset. The full combination of Charbonnier, VGG, and Sobel losses (blue line) achieves the highest SSIM score of 0.865, clearly outperforming all ablations. This demonstrates the complementary strengths of each loss: Charbonnier contributes to structural accuracy, VGG to perceptual consistency, and Sobel to edge preservation. Ablating any component results in noticeable drops, particularly the removal of Charbonnier or VGG, which significantly impairs SSIM convergence. Simpler loss terms like $L_1$, $L_2$, and LPIPS perform substantially worse throughout training, highlighting the importance of combining both distortion-based and perceptual components for optimal SR quality. Notably, our approach outperforms RLFN in both quality and training efficiency. Unlike RLFN, which employs a two-stage training procedure with $L_1$ and contrastive losses, we train our network end-to-end, achieving superior visual quality while reducing training time by 16\%.

\begin{wraptable}{r}{0.5\textwidth}
\caption{Ablation Study on Activation functions and Attention modules for EfRLFN. The best results appear in \textbf{bold}. $S(x)$ denotes a sigmoid function.}
\centering
\scriptsize
\setlength{\tabcolsep}{3.5pt}
\begin{tabular}{@{}llcccc@{}}
\toprule
\multicolumn{2}{c}{\textbf{Components}} & \multicolumn{4}{c}{\textbf{Metrics}} \\
\cmidrule(r){1-2} \cmidrule(l){3-6}
\textbf{Activation} & \textbf{Attention} & \textbf{SSIM} $\uparrow$ & \textbf{LPIPS} $\downarrow$ & \textbf{FPS} $\uparrow$ & \textbf{Params} $\downarrow$ \\
\midrule
\rowcolor{blue!10}
$tanh$ & ECA & \textbf{0.865} & 0.173 & \textbf{314} & \textbf{0.37M} \\
$tanh$ & ESA & 0.863 & \textbf{0.171} & 234 & 0.4M \\
$S(x) - 0.5$ & ECA & 0.856 & 0.179 & 305 & 0.37M \\
$S(x) - 0.5$ & ESA & 0.852 & 0.181 & 237 & 0.4M \\
$ReLU$ & ECA & 0.847 & 0.184 & 303 & 0.37M \\
$ReLU$ & ESA & 0.849 & 0.182 & 235 & 0.4M \\
\bottomrule
\end{tabular}
\label{tab:arch_ablation}
\end{wraptable}

\noindent \textbf{Architectural Analysis}  \:\:
Table~\ref{tab:arch_ablation} presents an ablation study isolating the effects of activation functions and attention mechanisms in EfRLFN. The combination of hyperbolic tangent activation and ECA attention yields the best overall performance, achieving the highest SSIM, strong perceptual quality, and the fastest runtime with a minimal parameter count. Replacing $tanh$ with a shifted sigmoid noticeably degrades SSIM and LPIPS across both attention variants. Similarly, using the heavier ESA attention block leads to lower frame rates and marginal gains in LPIPS at best. These results indicate that $tanh$ activation paired with the lightweight ECA module provides an optimal balance between quality and efficiency for real-time super-resolution.

Figure \ref{fig:preference}(b) shows output features from each ERLFB block of a model trained with different activation functions. Note that non-odd activation functions (ReLU) result in poor feature quality. Compared to $Sigmoid-0.5$, $tanh$ activation preserves more high-frequency features, which correlates with better detail restoration. This results in more detailed and cleaner outputs from the EfRLFN model.

\subsection{ONNX Runtime results}
To complement the PyTorch benchmarks, we exported EfRLFN, RLFN, and SPAN to ONNX and evaluated them using ONNX Runtime (v1.19.2) with both CUDA and TensorRT (v10.4.0) execution providers. All experiments were conducted on an NVIDIA RTX A6000 GPU using CUDA 11.7. The models were exported and executed in FP16 precision, and latency was measured on 360×480 inputs for both 2× and 4× upscaling.

As shown in Table \ref{tab:onnx}, EfRLFN consistently achieves lower latency than RLFN across both execution providers and surpasses real-time throughput ($\ge30$ FPS). The model also benefits significantly from TensorRT optimizations, further demonstrating its suitability for deployment-oriented inference engines.

\begin{table}[h]
\centering
\begin{tabular}{lccccc}
\hline
\textbf{Model} & \textbf{CUDA (ms)} & \textbf{TensorRT (ms)} & \textbf{Speedup} & \textbf{FPS (TRT)} & \textbf{Subjective Score} \\
\hline
\multicolumn{6}{c}{$2\times$ results} \\
\hline
RLFN            & 30.76 & 29.12 & 1.06$\times$ & 34.3 & 2.69 $\pm$ 0.15 \\
SPAN            & 16.29 & 10.46 & 1.55$\times$ & 95.6 & 2.55 $\pm$ 0.12 \\
EfRLFN (ours)   & 17.43 & 12.07 & 1.44$\times$ & 82.8 & \textbf{3.33 $\pm$ 0.14} \\
\hline
\multicolumn{6}{c}{$4\times$ results} \\
\hline
RLFN            & 38.24 & 32.48 & 1.18$\times$ & 30.8 & 4.32 $\pm$ 0.13 \\
SPAN            & 17.20 & 10.86 & 1.58$\times$ & 92.1 & 3.14 $\pm$ 0.12 \\
EfRLFN (ours)   & 20.25 & 14.66 & 1.38$\times$ & 68.2 & \textbf{4.52 $\pm$ 0.13} \\
\hline
\end{tabular}
\caption{ONNX Runtime results for $2\times$ and $4\times$ upscaling. FPS is measured with TensorRT.}
\label{tab:onnx}
\end{table}

\section{Conclusion}




Our research advances real-time super-resolution by introducing a comprehensive multi-scale dataset of 5,200 compressed videos, specifically designed for streaming applications. This dataset, coupled with systematic benchmarking of 11 SR models, provides valuable performance insights and establishes a robust evaluation framework. All resources - including the benchmark, dataset, and raw subjective comparison data (37,184 responses) - are publicly available at \url{https://github.com/EvgeneyBogatyrev/EfRLFN}. These subjective evaluations can serve as valuable data for training video quality assessment methods.

We further propose EfRLFN, an efficient SR model with novel architectural innovations and a composite loss training strategy, supporting both video and image inputs. Extensive evaluation demonstrates EfRLFN’s superior performance, delivering quality improvements and practical efficiency gains over existing methods. 
By combining carefully designed datasets, comprehensive benchmarking insights, and optimized model architecture, our work provides a complete foundation for advancing real-time SR research. Since EfRLFN is designed for single-image SR, exploring lightweight temporal extensions represents an important direction for future work. The released resources, including raw subjective data, enable future work in both super-resolution and perceptual quality assessment. 

\section{Acknowledgements}
This work was supported by the The Ministry of Economic Development of the Russian Federation in accordance with the subsidy agreement (agreement identifier 000000C313925P4H0002; grant No 139-15-2025-012).

The research was carried out using the MSU-270 supercomputer of Lomonosov Moscow State University.

\section{Reproducibility Statement}
To ensure the reproducibility of our work, we have made the following resources available. The complete source code and pre-trained weights for the proposed EfRLFN model are included in the supplementary materials. A detailed description of the video collection and preprocessing pipeline from the dataset is provided in Section \ref{sec:dataset}. Furthermore, the full set of experimental results, including additional ablations, can be found in Section \ref{app:a10-full-results}.



\bibliographystyle{iclr2026_conference}

\section{Appendix Contents}
This appendix provides additional details, experiments, and analyses to support the main paper. Below is a summary of each section:

\begin{itemize}
    \item \textbf{Section \ref{app:a1-limitations}}: Discussion of use-cases, limitations, and potential future improvements.
    \item \textbf{Section \ref{app:a2-real-time-sr}}: Summary of real-time super-resolution (SR) models evaluated, including architectures and training configurations.
    \item \textbf{Section \ref{app:a3-experimental-setup}}: Comprehensive description of the experimental setup.
    \item \textbf{Section \ref{app:a4-lr-hr-impact}}: Analysis of video pairs sourced from YouTube in comparison to bicubic downscaled LR.
    \item \textbf{Section \ref{app:a5-crowcourcing}}: Full methodology and results of the crowd-sourced perceptual evaluation.
    \item \textbf{Section \ref{sec:metric_corrs}}: Ablation study on visual quality metrics and their impact on evaluations.
    \item \textbf{Section \ref{sec:tanh_improvement}}: Ablation study on activation function replacements in the SPAN model.
    \item \textbf{Section \ref{sec:LLM}}: Prompts and sampled queries used for dataset collection.
    \item \textbf{Section \ref{app:a9-visual}}: Extended visual comparisons of SR model outputs.
    \item \textbf{Section \ref{app:a10-full-results}}: Expanded benchmark results (4$\times$ upscaling, additional metrics/datasets).
    \item \textbf{Section \ref{app:a12-dataset-preview}}: Dataset visualization and download links for the proposed StreamSR dataset.
\end{itemize}

\section{Discussion and Future Work}
\label{app:a1-limitations}

\paragraph{Computational Trade-offs.} 
Our benchmark intentionally excludes non-real-time super-resolution methods due to their low FPS. As visualized in Figure~\ref{fig:scatter}, state-of-the-art methods like RealESRGAN~\citep{wang2021real} and SwinIR~\citep{liang2021swinir} achieve marginally better perceptual quality but fall far below the 30 FPS threshold required for real-time applications on consumer GPUs (NVIDIA RTX 2080 GPU, 720p to 1440p upscaling). For instance, Omni-SR, the fastest non-real-time method in our comparison, operates at 16 FPS - significantly slower than SPAN, which at 60 FPS was the slowest real-time method we tested.
\paragraph{Codec Coverage Constraints.} 
The study's focus on YouTube-sourced video content naturally limits our evaluation to VP9, H.264, and AV1 codecs. While these account for most of web video traffic, specialized codecs used in professional settings (e.g., VVC) may exhibit different compression artifacts that could impact super-resolution performance. 

\paragraph{Societal Impact and Potential Misuses.}
Like all enhancement technologies, our model carries dual-use risks, such as unauthorized upscaling of low-resolution surveillance footage beyond legal limits or generating synthetic high-resolution content from deliberately degraded sources.  
\paragraph{Future Work.}
The real-time super-resolution technology offers significant benefits, including high-resolution streaming with lower bandwidth, which is particularly valuable for developing regions, but it also poses risks, where degraded content is enhanced to appear original. To support reproducibility and future research, we provide the complete implementation code, pretrained weights, validation datasets, and raw subjective evaluation scores as supplementary materials - this comprehensive package will enable other researchers to validate our findings, develop improved quality assessment methods, and advance real-time super-resolution techniques. Furthermore, the proposed StreamSR dataset and preference scores can be used to enhance existing super-resolution quality assessment techniques.

\section{Real-time SR Models Summary}
\label{app:a2-real-time-sr}

This section summarizes the real-time super-resolution models evaluated in our study. Table \ref{tab:model_specs} provides a comparative analysis of their key specifications, including model size (parameters), computational complexity (FLOPs), supported upscaling factors, architectural design, and training datasets.

\begin{table*}[ht!]
\caption{Model Specifications of Real-Time SR Methods. Architectural details and scaling capabilities are shown.}
\centering
\scriptsize
\setlength{\tabcolsep}{3.5pt} 
\begin{tabular}{lcccccc}
\toprule
\textbf{Method} & \textbf{Params (M)} & \textbf{FLOPs (G)} & \textbf{Upscale Factors} & \textbf{Architecture Type} & \textbf{Training Data} \\
\midrule
ESPCN~\citep{talab2019super} & 0.04 & 2.43 & 2$\times$,3$\times$,4$\times$ & Sub-pixel CNN & DIV2K \\
XLSR~\citep{ayazoglu2021extremely} & 0.03 & 1.81 & 4$\times$ & CNN & DIV2K \\
RLFN~\citep{kong2022residual} & 0.40 & 136 & 2$\times$,4$\times$ & CNN+Attention & DIV2K \\
AsConvSR~\citep{guo2023asconvsr} & 2.35 & 9 & 2$\times$ & Assembled Convolution & DIV2K+Flickr2K \\
Bicubic++~\citep{bilecen2023bicubic++} & 0.05 & 0.83  & 3$\times$ & CNN & DIV2K \\
NVIDIA VSR~\citep{nvidia-vsr}  & - & - & 2$\times$,3$\times$,4$\times$ & CNN+Attention & Proprietary \\
RT4KSR~\citep{zamfir2023towards} & 0.05 & 172 & 2$\times$ & CNN+Attention & DIV2K+Flickr2K \\
SAFMN~\citep{sun2023safmn} & 0.23 & 52 & 4$\times$ & CNN with Feature modulation & DIV2K+Flickr2K \\
SMFANet~\citep{smfanet} & 0.20 & 11 & 4$\times$ & CNN with Feature modulation & DIV2K+Flickr2K \\
SPAN~\citep{wan2024swift} & 0.43 & 28 & 2$\times$,4$\times$ & Parameter-free attention & DIV2K+Flickr2K \\
\bottomrule
\end{tabular}
\label{tab:model_specs}
\end{table*}

\begin{table}[h!]
\caption{Performance Comparison of Training Strategies on DIV2K.}
\centering
\scriptsize
\begin{tabular}{@{}lccccc@{}}
\toprule
\textbf{Model} & \textbf{LR Type} & \textbf{PSNR} $\uparrow$ & \textbf{SSIM} $\uparrow$ & \textbf{LPIPS} $\downarrow$ & \textbf{ERQA} $\uparrow$ \\
\midrule
\multirow{2}{*}{SPAN} 
    & Synthetic & 32.645 & 0.834 & 0.227 & 0.489 \\
    & \cellcolor{blue!10}Real LR & \cellcolor{blue!10}\textbf{33.511} & \cellcolor{blue!10}\textbf{0.847} & \cellcolor{blue!10}\textbf{0.206} & \cellcolor{blue!10}\textbf{0.501} \\
    
\addlinespace[0.2em]
\multirow{2}{*}{RLFN} 
    & Synthetic & 32.718 & 0.845 & 0.193 & 0.517 \\
    & \cellcolor{blue!10}Real LR & \cellcolor{blue!10}\textbf{33.816} & \cellcolor{blue!10}\textbf{0.855} & \cellcolor{blue!10}\textbf{0.182} & \cellcolor{blue!10}\textbf{0.535} \\
    
\addlinespace[0.2em]
\multirow{2}{*}{EfRLFN} 
    & Synthetic & 33.983 & 0.859 & 0.179 & 0.529 \\
    & \cellcolor{blue!10}Real LR & \cellcolor{blue!10}\textbf{34.553} & \cellcolor{blue!10}\textbf{0.865} & \cellcolor{blue!10}\textbf{0.173} & \cellcolor{blue!10}\textbf{0.536} \\
\bottomrule
\end{tabular}
\label{tab:bicubic-vs-real}
\end{table}

\section{Experimental Setup}
\label{app:a3-experimental-setup}

We trained each model for 500 epochs, following the training processes described in the original papers. All models were evaluated on 720p videos with 2$\times$ upscaling to 1440p resolution. For FPS measurements, we ran each model on an NVIDIA GeForce RTX 2080 GPU and calculated runtime on the same test sequence of 100 frames, averaged across 3 sequential runs. During EfRLFN training we set loss coefficients $\lambda_\text{Charb} = 1$, $\lambda_\text{VGG} = 10^{-3}$, and $\lambda_\text{Sobel} = 10^{-1}$ for normalization purposes.

\section{The Impact of Using LR-HR Pairs}
\label{app:a4-lr-hr-impact}

SPAN, RLFN, and EfRLFN models trained using HR-LR pairs from our dataset show better performance (Table \ref{tab:bicubic-vs-real}) compared to the same models trained using a bicubic downsampling process. In the latter case, the HR images from our dataset were downsampled to generate LR images using bicubic interpolation. This observation highlights the effectiveness of collecting domain-specific HR-LR pairs directly from YouTube, demonstrating that these pairs are more representative and beneficial than synthetic bicubic downsampling.

\section{Crowd-Sourced Study Details}
\label{app:a5-crowcourcing}

The evaluation was organized into two distinct tracks: one focusing on 2$\times$ SR and the other on 4$\times$ SR. Given the limited number of assessors with screen resolutions exceeding Full HD, we opted to present each test video as a center Full HD crop and showed videos sequentially.

During the experiment, participants were presented with pairs of videos generated by two randomly selected SR models. They were tasked with selecting the video that exhibited fewer visual artifacts; an option to indicate that the videos were “indistinguishable” was also provided. Each pair of videos was assessed by a total of 30 participants, with each participant evaluating 10 pairs of videos.

To ensure the integrity of the results, we included two verification questions with defined correct answers among the 10 pairs. Responses from any participant who failed to answer one or more of these verification questions correctly were excluded from further analysis. A total of 3,822 unique individuals participated in our subjective evaluation. The final subjective scores were derived from the 37,184 responses, applying the Bradley-Terry model~\citep{bradley1952rank}.

\begin{figure*}[h!]
\centering
\includegraphics[width=0.99\linewidth]{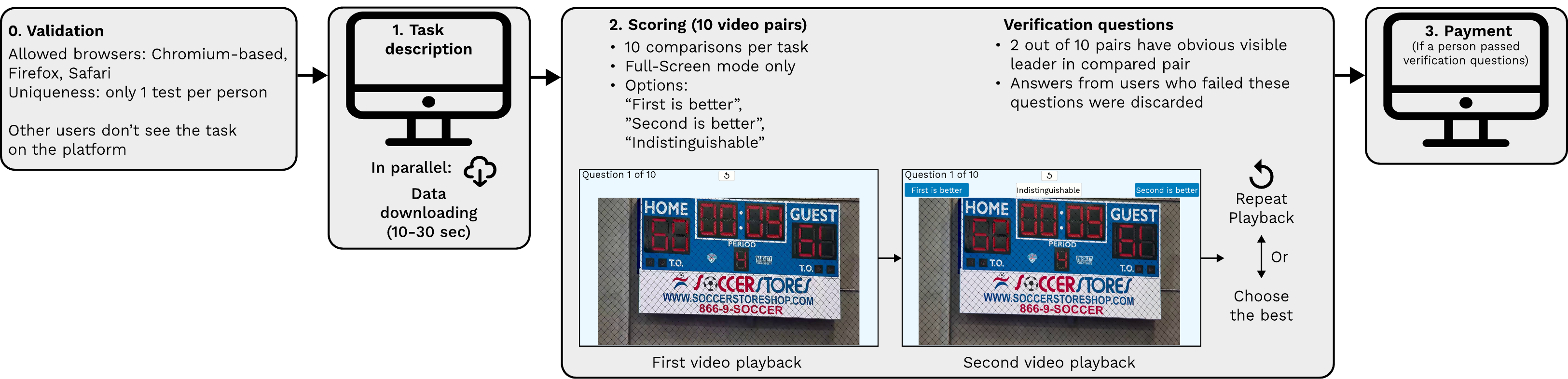}
\caption{An overall scheme of the crowdsource subjective study described in Section \ref{app:a5-crowcourcing}. A screenshot of the interface of crowdsource subjective platform.}
\label{fig:subjectify-interface}
\end{figure*}

Figure \ref{fig:subjectify-interface} summarizes our setup for the study and visualizes the interface of Subjectify.us service used for crowd-sourcing.

Before completing the comparison, each participant was shown the following instructions:

\begin{quote}
You will see video pairs shown one after another. For each pair, select the video with better visual quality (sharper, with fewer distortions like blurring or pixelation). If they look identical, choose "Indistinguishable."

Be careful—there are verification questions in this test. Incorrect answers on verification checks will lead to rejection.
\end{quote}

\section{Correlation of Metrics with Subjective Scores}
\label{sec:metric_corrs}

\begin{figure}[h!]
\centering
\includegraphics[width=0.8\linewidth]{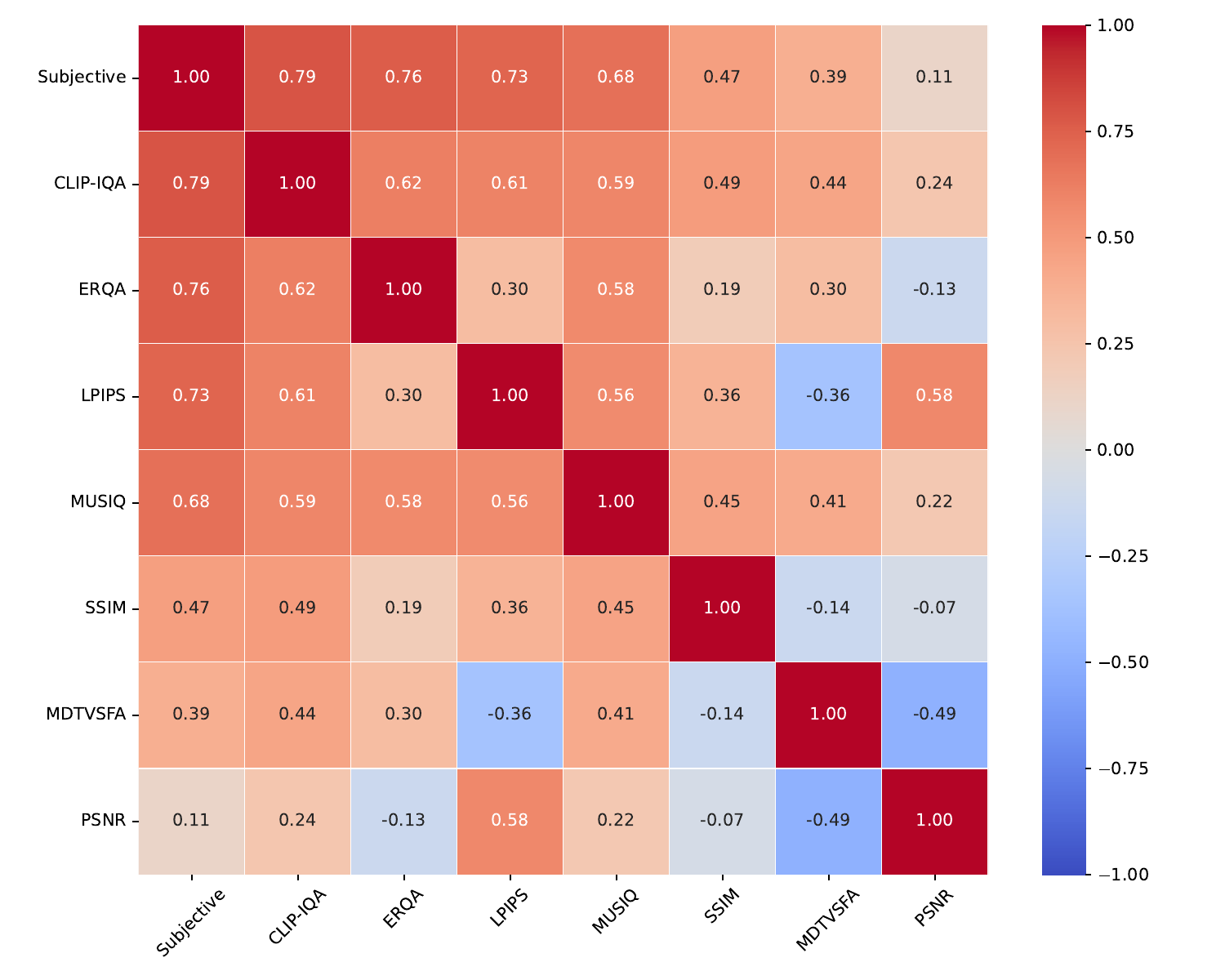}
\caption{Pearson Correlation between objective quality metrics and subjective scores.}
\label{fig:metrics-square}
\end{figure}

Figure~\ref{fig:metrics-square} presents the correlation analysis between objective video quality metrics and human subjective evaluations. Our experimental results demonstrate that four metrics -- ClipIQA, ERQA, LPIPS, and MUSIQ -- show significant correlations with subjective quality assessments. 


These findings align with prior research in three key aspects: (1) the superior performance of feature-based metrics (LPIPS, MUSIQ) over traditional approaches, (2) ERQA's specialization for restoration quality assessment, and (3) the poor correlation of PSNR with human perception, consistent with other works \citep{ huynh2008scope, bogatyrev2023srcodec}.

\begin{figure*}[h!]
\centering
\includegraphics[width=0.9\linewidth]{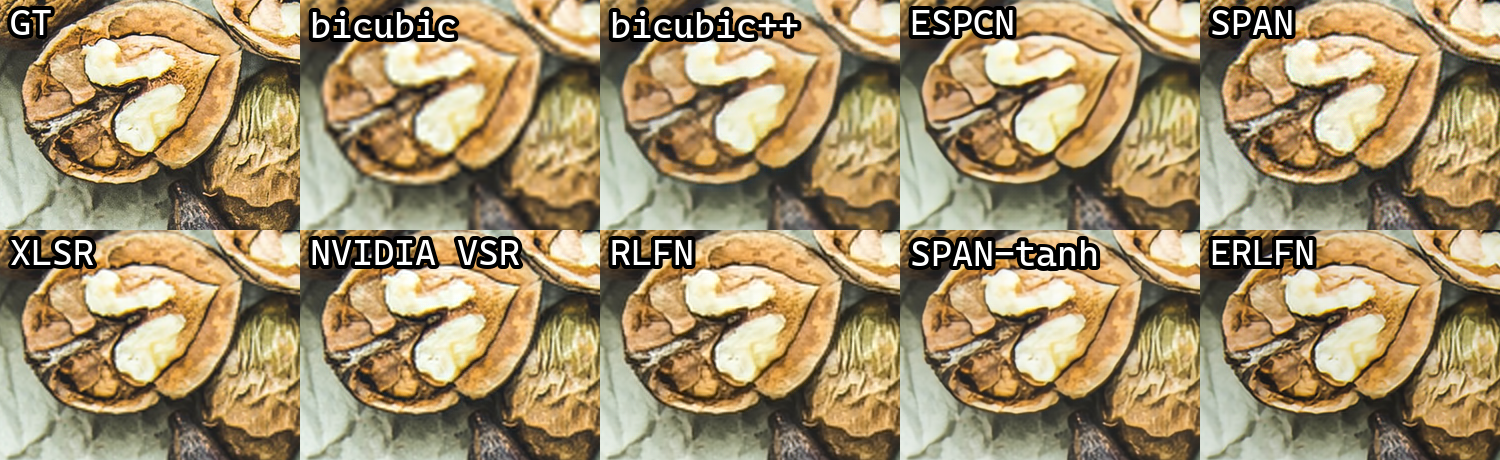}
\includegraphics[width=0.9\linewidth]{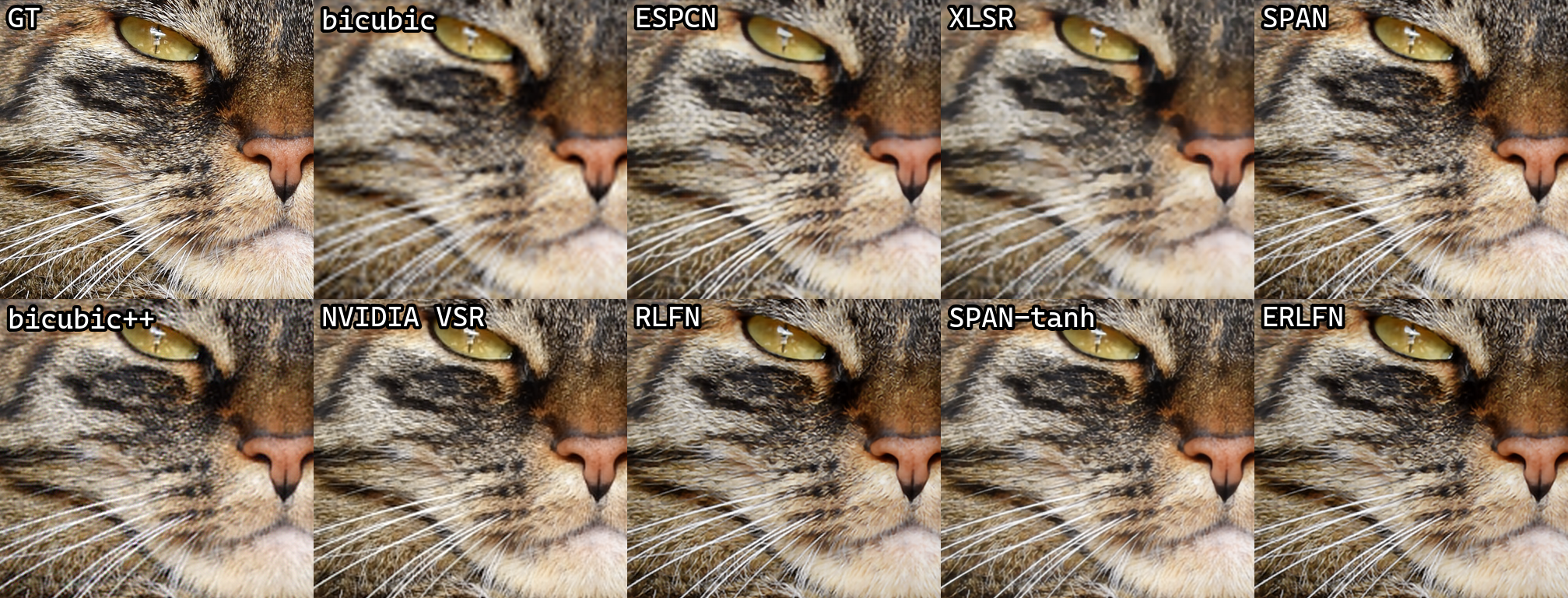}
\caption{The comparison between several 2$\times$ real-time SR models.}
\label{fig:real-time-sr-compare}
\end{figure*}


\section{Improving SPAN with Hyperbolic Tangent Activation}
\label{sec:tanh_improvement}

Our experiments demonstrate that replacing the shifted sigmoid ($\text{Sigmoid}(x)-0.5$) with hyperbolic tangent ($\tanh$) activation in SPAN leads to consistent improvements over the original formulation. Table~\ref{tab:tanh_vs_original} compares both variants across standard benchmarks:

\begin{table}[h]
\caption{Performance Comparison: SPAN ($\sigma=0.5$) vs. SPAN ($\tanh$).}
\centering
\scriptsize
\begin{tabular}{@{}lcccc@{}}
\toprule
 & \textbf{Set14} & \textbf{BSD100} & \textbf{Urban100} & \textbf{DIV2K} \\
\textbf{Model} & SSIM$\uparrow$/LPIPS$\downarrow$ & SSIM$\uparrow$/LPIPS$\downarrow$ & SSIM$\uparrow$/LPIPS$\downarrow$ & SSIM$\uparrow$/LPIPS$\downarrow$ \\
\midrule
SPAN ($\sigma=0.5$) & 0.836 / 0.142 & 0.837 / 0.239 & 0.847 / 0.118 & 0.890 / 0.175 \\
\rowcolor{blue!10}
SPAN ($\tanh$) & \textbf{0.837} / \textbf{0.140} & \textbf{0.841} / \textbf{0.228} & \textbf{0.853} / \textbf{0.115} & \textbf{0.891} / \textbf{0.167} \\
\bottomrule
\end{tabular}
\label{tab:tanh_vs_original}
\end{table}

\section{LLM-generated Queries}
\label{sec:LLM}

Table~\ref{tab:youtube_categories} presents the description of video categories used in our study, including their descriptions and the corresponding prompt templates used for query generation. Table~\ref{tab:video_categories} complements this by providing representative examples of LLM-generated search queries organized by these categories, demonstrating the diversity of content collected for our benchmark dataset.

\begin{figure*}[h!]
\begin{minipage}[t]{0.48\linewidth}
    \centering
    \includegraphics[width=\linewidth]{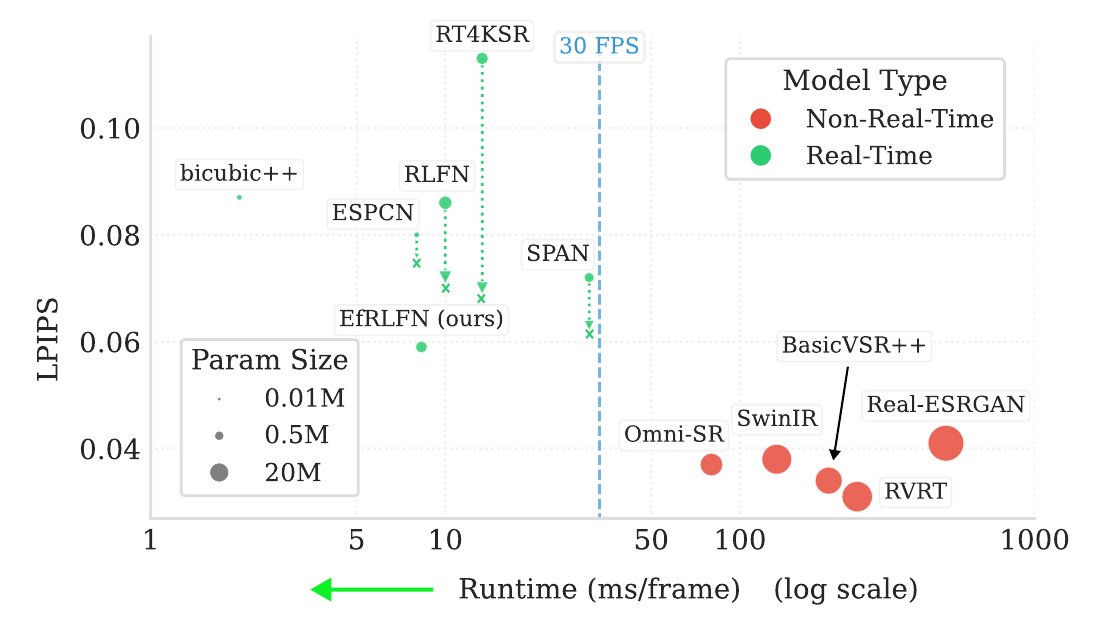}
    \caption{Trade-off between LPIPS and run-time speed for various 2$\times$ super-resolution models.}
    \label{fig:lpips-to-runtime-scatter}
\end{minipage}
\hfill
\begin{minipage}[t]{0.48\linewidth}
    \centering
    \includegraphics[width=\linewidth]{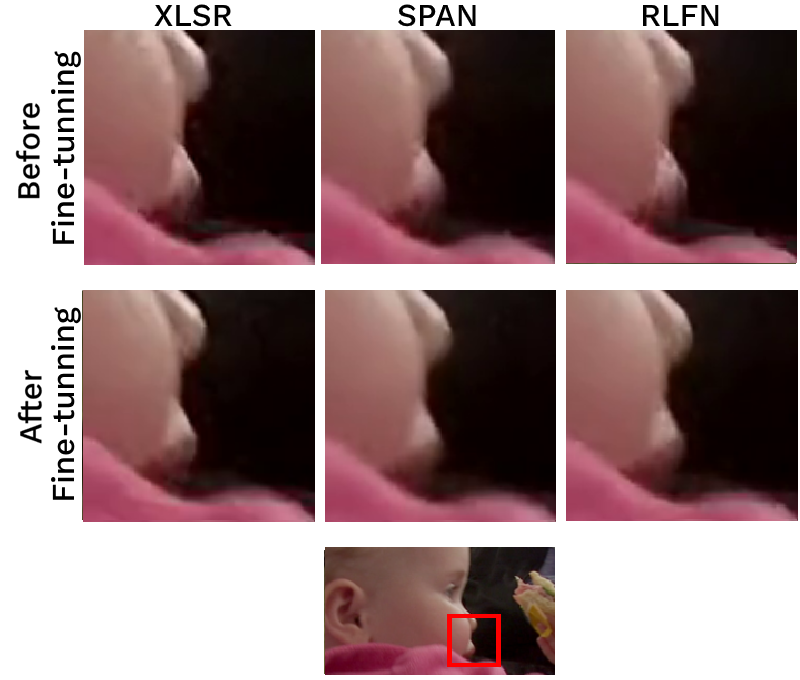}
    \caption{Visual quality comparison before and after fine-tuning on our dataset.}
    \label{fig:visual-example-appendix}
\end{minipage}
\end{figure*}

\begin{figure*}[h!]
\begin{minipage}[c]{0.495\linewidth}
  \centering
  \includegraphics[width=1.0\linewidth]{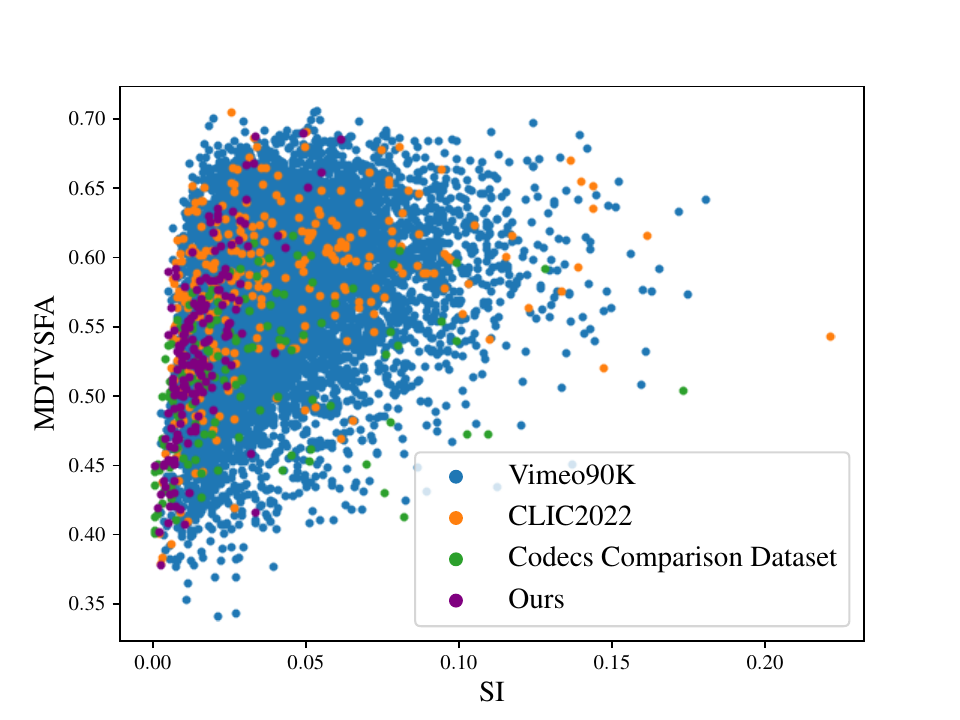}
\end{minipage}%
\begin{minipage}[c]{0.495\linewidth}
  \centering
  \includegraphics[width=1.0\linewidth]{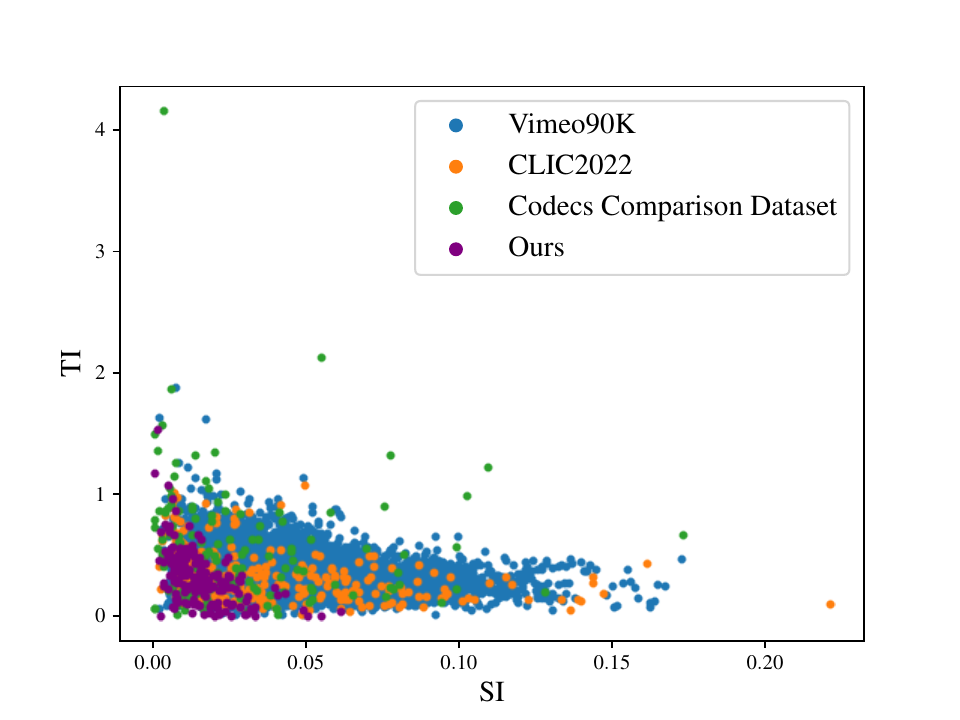}
\end{minipage}%
\caption{Distribution of Google Spatial and Temporal~\citep{wang2019youtube} information and MDTVSFA~\citep{li2021unified} for videos from different datasets.}
\label{fig:si-ti-mdtvsfa}
\end{figure*}

\begin{table*}[th!]
\centering
\caption{LLM-generated queries examples.}
\begin{tabular}{@{}p{4.5cm}p{4.5cm}p{4.5cm}@{}}
\toprule
\textbf{Sports} & \textbf{Animals} & \textbf{Human Activities} \\
\begin{enumerate}[itemsep=0pt, topsep=2pt, leftmargin=*, nosep]
    \item "highlights of football"
    \item "sports slow motion"
    \item "extreme sports action"
    \item "basketball dunks"
    \item "soccer goals"
\end{enumerate} &
\begin{enumerate}[itemsep=0pt, topsep=2pt, leftmargin=*, nosep]
    \item "wildlife documentary"
    \item "cute animal videos"
    \item "slow motion animals"
    \item "pets playing"
    \item "nature interactions"
\end{enumerate} &
\begin{enumerate}[itemsep=0pt, topsep=2pt, leftmargin=*, nosep]
    \item "people dancing"
    \item "cooking tutorials"
    \item "fitness workouts"
    \item "street performers"
    \item "DIY projects"
\end{enumerate} \\
\midrule
\textbf{Art and Crafts} & \textbf{Macro Videos} & \textbf{Historical Footage} \\
\begin{enumerate}[itemsep=0pt, topsep=2pt, leftmargin=*, nosep]
    \item "time-lapse painting"
    \item "sculpting techniques"
    \item "DIY crafts"
    \item "artistic creations"
    \item "watercolor tutorial"
\end{enumerate} &
\begin{enumerate}[itemsep=0pt, topsep=2pt, leftmargin=*, nosep]
    \item "macro insects"
    \item "close-up flowers"
    \item "macro techniques"
    \item "tiny world"
    \item "everyday objects"
\end{enumerate} &
\begin{enumerate}[itemsep=0pt, topsep=2pt, leftmargin=*, nosep]
    \item "historical footage"
    \item "old documentaries"
    \item "historical events"
    \item "classic film clips"
    \item "moon landing"
\end{enumerate} \\
\midrule
\textbf{Animated Content} & \textbf{Science} & \textbf{Nature/Landscapes} \\
\begin{enumerate}[itemsep=0pt, topsep=2pt, leftmargin=*, nosep]
    \item "animation films"
    \item "animated scenes"
    \item "cartoon clips"
    \item "stop motion"
    \item "3D animation"
\end{enumerate} &
\begin{enumerate}[itemsep=0pt, topsep=2pt, leftmargin=*, nosep]
    \item "science experiments"
    \item "physics demos"
    \item "chemical reactions"
    \item "biology explained"
    \item "science phenomena"
\end{enumerate} &
\begin{enumerate}[itemsep=0pt, topsep=2pt, leftmargin=*, nosep]
    \item "nature landscapes"
    \item "time-lapse nature"
    \item "ocean waves"
    \item "forest scenery"
    \item "mountain views"
\end{enumerate} \\
\bottomrule
\end{tabular}
\label{tab:video_categories}
\end{table*}

\begin{table*}[ht]
\centering
\scriptsize
\caption{YouTube Content Categories and Query Generation Prompts.}
\begin{tabular}{@{}ccc@{}}
\toprule
\textbf{Category Name} & \textbf{Category Definition} & \textbf{Query Generation Prompt} \\
\midrule
Nature and Landscapes & Scenic natural environments and geographical features. & \makecell{"Generate 100 diverse YouTube search queries about nature and landscapes, \\ including scenic views, wildlife habitats, waterfalls, forests, and time-lapse \\ nature videos."} \\
\addlinespace

Urban Environments & Human-made structures and city landscapes. & \makecell{"Provide 100 YouTube search queries related to urban environments, \\ such as cityscapes, street photography, architectural marvels, urban \\ exploration, and drone footage of cities."} \\
\addlinespace

Sports & Athletic activities and competitive physical games. & \makecell{"List 100 YouTube search queries for sports content, including game \\ highlights, athlete interviews, training routines, extreme sports, and \\ sports analysis."} \\
\addlinespace

Animals & Living organisms from the animal kingdom. & \makecell{"Suggest 100 YouTube search queries about animals, covering wildlife \\ documentaries, pet care tips, rare animal sightings, funny animal \\ compilations, and animal behavior studies."} \\
\addlinespace

Human Activities & Actions and behaviors performed by people. & \makecell{"Generate 100 YouTube search queries focused on human activities, \\ such as daily routines, cultural festivals, street performances, social \\ experiments, and work-life documentaries."} \\
\addlinespace

Art and Crafts & Creative works and handmade objects. & \makecell{"Create 100 YouTube search queries for art and crafts, including painting \\ tutorials, DIY projects, sculpture making, digital art timelapses, and craft \\ ideas for beginners."} \\
\addlinespace

Macro Videos & Extreme close-up footage of small subjects. & \makecell{"Provide 100 YouTube search queries for macro videos, such as close-up \\ insect footage, microscopic water droplets, detailed textures, and extreme \\ close-up photography."} \\
\addlinespace

Historical Footage & Recorded material from past events. & \makecell{"List 100 YouTube search queries for historical footage, including old war \\ videos, vintage city life, archival documentaries, and restored historical \\ clips."} \\
\addlinespace

Animated Content & Moving images created through animation techniques. & \makecell{"Suggest 100 YouTube search queries for animated content, like short \\ animated films, 3D animation breakdowns, cartoon compilations, and motion \\ graphics tutorials."} \\
\addlinespace

Science & Illustrations of scientific principles. & \makecell{"Generate 100 YouTube search queries about scientific demonstrations, \\ including chemistry experiments, physics simulations, biology dissections, \\ and engineering prototypes."} \\
\addlinespace

Technology Reviews & Evaluations of technological products. & \makecell{"Provide 100 YouTube search queries for technology reviews, such as \\ smartphone comparisons, gadget unboxings, software tutorials, and \\ futuristic tech showcases."} \\
\addlinespace

Cooking and Recipes & Food preparation instructions and techniques. & \makecell{"List 100 YouTube search queries for cooking and recipes, including easy \\ meal prep, gourmet dishes, baking tutorials, street food tours, and kitchen \\ hacks."} \\
\addlinespace

Travel Vlogs & Video blogs documenting travel experiences. & \makecell{"Suggest 100 YouTube search queries for travel vlogs, covering backpacking \\ adventures, luxury travel guides, hidden tourist spots, and cultural \\ immersion experiences."} \\
\addlinespace

Fitness and Workouts & Physical exercise routines and training. & \makecell{"Generate 100 YouTube search queries for fitness and workouts, like home \\ exercise routines, gym training tips, yoga sessions, and body \\ transformation stories."} \\
\addlinespace

Music Performances & Live or recorded musical presentations. & \makecell{"Provide 100 YouTube search queries for music performances, including \\ live concerts, acoustic covers, music festival highlights, and street \\ musician videos."} \\
\addlinespace

Educational Tutorials & Instructional content for learning. & \makecell{"List 100 YouTube search queries for educational tutorials, such as math \\ problem-solving, language learning, coding lessons, and science \\ explainers."} \\
\addlinespace

Gaming Content & Video game-related material. & \makecell{"Suggest 100 YouTube search queries for gaming content, like walkthroughs, \\ esports tournaments, game reviews, speedruns, and funny gaming \\ moments."} \\
\addlinespace

DIY Projects & Do-it-yourself creative endeavors. & \makecell{"Generate 100 YouTube search queries for DIY projects, including home \\ improvement hacks, handmade crafts, upcycling ideas, and woodworking \\ tutorials."} \\
\addlinespace

Drone Footage & Aerial video captured by drones. & \makecell{"Provide 100 YouTube search queries for drone footage, such as aerial nature \\ shots, city flyovers, mountain landscapes, and cinematic drone \\ cinematography."} \\
\addlinespace

Time-lapse Videos & Accelerated video sequences showing change over time. & \makecell{"List 100 YouTube search queries for time-lapse videos, including sunset \\ transitions, city day-to-night changes, plant growth, and construction \\ progressions."} \\
\bottomrule
\end{tabular}
\label{tab:youtube_categories}
\end{table*}

\section{Visual Comparison}
\label{app:a9-visual}
Figure~\ref{fig:real-time-sr-compare} compares real-time SR models on DIV2K. Figure~\ref{fig:visual-example-appendix} demonstrates significant quality gains after StreamSR fine-tuning, particularly in edge preservation and temporal stability.

\section{Full Results}
\label{app:a10-full-results}
Table~\ref{tab:full_results_rotate} presents comprehensive full results of our benchmark. Tables~\ref{tab:2x-results-app} and~\ref{tab:4x-results-app} provide full validation results on other datasets, including Set14~\citep{zeyde2012single} and REDS~\citep{nah2019ntire}.




\begin{sidewaystable}[t!]
\caption{Comparison of real-time SR models on popular benchmarks (2× upscaling). Best results in \textbf{bold}, second best \underline{underlined}. ``$^T$'' indicates fine-tuning on our dataset.}
\centering
\scriptsize
\begin{tabular}{@{}lcccccc@{}}
\toprule
\textbf{Model} & \textbf{BSD100} & \textbf{Urban100} & \textbf{Set14} & \textbf{REDS} & \textbf{DIV2K} \\
 & \small PSNR$\uparrow$/SSIM$\uparrow$/LPIPS$\downarrow$ & \small PSNR$\uparrow$/SSIM$\uparrow$/LPIPS$\downarrow$ & \small PSNR$\uparrow$/SSIM$\uparrow$/LPIPS$\downarrow$ & \small PSNR$\uparrow$/SSIM$\uparrow$/LPIPS$\downarrow$ & \small PSNR$\uparrow$/SSIM$\uparrow$/LPIPS$\downarrow$ & \\
\midrule
RT4KSR & 28.1/0.805/0.296 & 27.3/0.769/0.244 & 28.9/0.782/0.231 & 26.7/0.758/0.268 & 30.2/0.860/0.221 \\
RT4KSR$^{T}$ & 28.3/0.812/0.254 & 27.5/0.766/0.223 & 29.1/0.789/0.210 & 26.9/0.765/0.245 & 30.5/0.864/0.186 \\
AsConvSR$^{T}$ & 28.7/0.832/0.270 & 28.1/0.824/0.208 & 29.5/0.801/0.205 & 27.3/0.781/0.238 & 30.8/0.883/0.206 \\
bicubic & 27.0/0.752/0.386 & 26.2/0.711/0.324 & 27.8/0.725/0.308 & 25.6/0.702/0.342 & 29.1/0.820/0.292  \\
bicubic++$^{T}$ & 27.3/0.768/0.360 & 26.5/0.725/0.298 & 28.1/0.741/0.285 & 25.9/0.718/0.318 & 29.4/0.832/0.270\\
ESPCN & 24.5/0.492/0.534 & 23.8/0.505/0.437 & 25.2/0.518/0.432 & 23.4/0.487/0.478 & 26.1/0.514/0.515\\
ESPCN$^{T}$ & 28.2/0.814/0.236 & 27.6/0.840/0.150 & 29.0/0.795/0.185 & 26.8/0.772/0.215 & 30.4/0.862/0.180 \\
XLSR & 27.9/0.788/0.243 & 27.2/0.786/0.152 & 28.7/0.772/0.178 & 26.4/0.751/0.208 & 30.1/0.858/0.180 \\
XLSR$^{T}$ & 28.4/0.817/0.244 & 27.8/0.828/0.158 & 29.2/0.801/0.183 & 26.9/0.779/0.216 & 30.5/0.864/0.183 \\
SMFANet$^{T}$ & 28.1/0.803/0.239 & 27.5/0.798/0.153 & 28.9/0.787/0.180 & 26.7/0.766/0.209 & 30.4/0.865/0.177 \\
SAFMN$^{T}$ & 28.4/0.813/0.236 & 27.8/0.808/0.150 & 29.2/0.797/0.178 & 27.0/0.777/0.207 & 30.7/0.871/0.175 \\
RLFN & 28.2/0.805/0.238 & 27.6/0.803/0.147 & 29.0/0.792/0.172 & 26.8/0.771/0.203 & 30.6/0.876/0.175 \\
RLFN$^{T}$ & 28.9/0.834/0.239 & 28.3/0.845/0.153 & 29.7/0.822/0.175 & 27.5/0.801/0.206 & 31.1/0.881/0.178  \\
SPAN & 28.9/0.836/\underline{0.222} & 28.4/0.841/0.139 & 29.8/0.819/0.155 & 27.6/0.798/0.186 & 31.2/0.887/\underline{0.168}  \\
SPAN$^{T}$ & \underline{29.0}/\underline{0.837}/0.239 & \underline{28.5}/\underline{0.847}/\underline{0.118} & \underline{29.9}/\underline{0.826}/0.170 & \underline{27.7}/\underline{0.805}/0.205 & \underline{31.3}/\underline{0.890}/0.175  \\
\rowcolor{blue!10}
\textbf{EfRLFN}$^{T}$ & \textbf{29.3}/\textbf{0.847}/\textbf{0.190} & \textbf{28.8}/\textbf{0.856}/\textbf{0.116} & \textbf{30.2}/\textbf{0.835}/\textbf{0.148} & \textbf{28.0}/\textbf{0.814}/\textbf{0.178} & \textbf{31.5}/\textbf{0.892}/\textbf{0.145} \\
\bottomrule
\end{tabular}
\label{tab:2x-results-app}
\end{sidewaystable}

\begin{sidewaystable}[t!]
\caption{Comparison of real-time SR models on popular benchmarks (4× upscaling). Best results in \textbf{bold}, second best \underline{underlined}. ``$^T$'' indicates fine-tuning on our dataset.}
\centering
\scriptsize
\begin{tabular}{@{}lcccccc@{}}
\toprule
\textbf{Model} & \textbf{BSD100} & \textbf{Urban100} & \textbf{Set14} & \textbf{REDS} & \textbf{DIV2K} \\
 & \small PSNR$\uparrow$/SSIM$\uparrow$/LPIPS$\downarrow$ & \small PSNR$\uparrow$/SSIM$\uparrow$/LPIPS$\downarrow$ & \small PSNR$\uparrow$/SSIM$\uparrow$/LPIPS$\downarrow$ & \small PSNR$\uparrow$/SSIM$\uparrow$/LPIPS$\downarrow$ & \small PSNR$\uparrow$/SSIM$\uparrow$/LPIPS$\downarrow$ & \\
\midrule
RT4KSR$^{T}$ & 24.8/0.409/0.575 & 23.7/0.489/0.521 & 25.1/0.452/0.503 & 23.5/0.471/0.542 & 27.3/0.696/0.489 \\
AsConvSR$^{T}$ & 25.1/0.433/0.586 & 24.0/0.500/0.512 & 25.4/0.468/0.488 & 23.8/0.489/0.518 & 27.6/0.701/0.471  \\
bicubic & 24.9/0.423/0.589 & 23.8/0.493/0.589 & 25.2/0.445/0.521 & 23.6/0.463/0.553 & 26.8/0.562/0.484 \\
ESPCN & 25.0/0.434/0.671 & 23.9/0.432/0.597 & 25.3/0.447/0.563 & 23.7/0.438/0.602 & 26.9/0.539/0.572  \\
ESPCN$^{T}$ & 25.2/0.442/0.662 & 24.1/0.452/0.585 & 25.5/0.459/0.548 & 23.9/0.445/0.587 & 27.1/0.551/0.557  \\
XLSR & 26.8/0.587/0.497 & 25.7/0.580/0.466 & 27.1/0.562/0.421 & 25.5/0.548/0.451 & 28.9/0.704/0.391 \\
XLSR$^{T}$ & 27.5/0.659/0.450 & 26.4/0.672/0.341 & 27.8/0.631/0.388 & 26.2/0.612/0.402 & 29.6/0.759/0.339  \\
SMFANet & 27.0/0.602/0.488 & 25.9/0.595/0.458 & 27.3/0.578/0.413 & 25.7/0.563/0.438 & 29.1/0.715/0.384 \\
SMFANet$^{T}$ & 27.2/0.612/0.482 & 26.1/0.608/0.450 & 27.5/0.588/0.407 & 25.9/0.573/0.432 & 29.3/0.722/0.380  \\
SAFMN & 27.3/0.622/0.475 & 26.2/0.620/0.442 & 27.6/0.601/0.402 & 26.0/0.586/0.425 & 29.4/0.729/0.376 \\
SAFMN$^{T}$ & 27.5/0.632/0.468 & 26.4/0.632/0.434 & 27.8/0.611/0.398 & 26.2/0.596/0.420 & 29.6/0.736/0.372 \\
RLFN & 27.1/0.603/0.492 & 26.0/0.597/0.460 & 27.4/0.578/0.409 & 25.8/0.563/0.434 & 29.2/0.721/0.386 \\
RLFN$^{T}$ & \underline{27.9}/\underline{0.674}/\underline{0.445} & \underline{26.8}/\underline{0.688}/\underline{0.336} & \underline{28.2}/\underline{0.647}/\underline{0.382} & \underline{26.6}/\underline{0.632}/\underline{0.398} & \underline{30.0}/\underline{0.774}/\underline{0.334} \\
SPAN & 27.5/0.632/0.481 & 26.4/0.639/0.369 & 27.8/0.618/0.358 & 26.2/0.603/0.384 & 29.6/0.739/0.341 \\
SPAN$^{T}$ & 27.7/0.649/0.483 & 26.6/0.650/0.406 & 28.0/0.625/0.375 & 26.4/0.610/0.401 & 29.8/0.750/0.375 \\
\rowcolor{blue!10}
\textbf{EfRLFN}$^{T}$ & \textbf{28.2}/\textbf{0.682}/\textbf{0.429} & \textbf{27.1}/\textbf{0.699}/\textbf{0.327} & \textbf{28.5}/\textbf{0.658}/\textbf{0.365} & \textbf{26.9}/\textbf{0.643}/\textbf{0.381} & \textbf{30.3}/\textbf{0.778}/\textbf{0.331} \\
\bottomrule
\end{tabular}
\label{tab:4x-results-app}
\end{sidewaystable}

\begin{sidewaystable}
\caption{Ranking of real-time SR methods with 95\% confidence intervals. Best results in \textbf{bold}. ``T'' indicates fine-tuning. N/A means that model does not have pretrained weights for a specific upscaling factor.}
\centering
\tiny
\begin{tabular}{lcccccccccccccccc}
\toprule
 & \multicolumn{8}{c}{\textbf{2$\times$ Track}} & \multicolumn{8}{c}{\textbf{4$\times$ Track}} \\
\cmidrule(lr){2-9} \cmidrule(lr){10-17}
\textbf{Method} & Subj. & PSNR & SSIM & LPIPS & MUSIQ & CLIP-IQA & ERQA & MDTVSFA & Subj. & PSNR & SSIM & LPIPS & MUSIQ & CLIP-IQA & ERQA & MDTVSFA \\
\midrule
AsConvSR$^{T}$ & 1.93±0.11 & 35.25±0.15 & 0.912±0.003 & 0.214±0.008 & 52.3±1.2 & 0.48±0.02 & 0.53±0.01 & 0.512±0.008 & 1.32±0.14 & 32.65±0.15 & 0.865±0.003 & 0.382±0.008 & 41.2±1.1 & 0.38±0.02 & 0.42±0.01 & 0.495±0.007 \\
RT4KSR & 2.40±0.12 & 36.45±0.16 & 0.918±0.003 & 0.213±0.009 & 54.1±1.1 & 0.49±0.02 & 0.54±0.01 & 0.518±0.008 & 1.40±0.14 & 32.65±0.16 & 0.866±0.003 & 0.382±0.009 & 39.8±1.0 & 0.36±0.02 & 0.41±0.01 & 0.492±0.007 \\
RT4KSR$^{T}$ & 2.43±0.11 & 37.55±0.14 & 0.925±0.002 & 0.070±0.003 & 58.7±1.3 & 0.53±0.02 & 0.58±0.01 & 0.527±0.008 & 1.45±0.14 & 32.65±0.14 & 0.867±0.002 & 0.382±0.003 & 42.1±1.3 & 0.39±0.02 & 0.43±0.01 & 0.498±0.007 \\
bicubic & 2.44±0.14 & 30.32±0.21 & 0.872±0.004 & 0.076±0.005 & 49.5±1.2 & 0.44±0.024 & 0.48±0.01 & 0.502±0.008 & 1.49±0.13 & 32.81±0.21 & 0.842±0.004 & 0.240±0.005 & 42.5±1.3 & 0.40±0.024 & 0.45±0.01 & 0.503±0.007 \\
ESPCN & 2.48±0.12 & 30.71±0.18 & 0.878±0.003 & 0.078±0.004 & 50.2±1.0 & 0.45±0.02 & 0.49±0.01 & 0.505±0.008 & 1.52±0.11 & 32.04±0.18 & 0.838±0.003 & 0.276±0.004 & 43.1±1.1 & 0.41±0.02 & 0.46±0.01 & 0.506±0.007 \\
ESPCN$^{T}$ & 2.49±0.11 & 35.76±0.15 & 0.908±0.002 & 0.072±0.003 & 56.9±1.2 & 0.51±0.02 & 0.55±0.01 & 0.521±0.008 & 1.96±0.15 & 32.08±0.15 & 0.848±0.002 & 0.215±0.003 & 45.3±1.2 & 0.44±0.02 & 0.48±0.01 & 0.512±0.007 \\ 
Bicubic++$^{T}$ & 2.44±0.13 & 36.79±0.15 & 0.915±0.002 & 0.087±0.004 & 57.2±1.3 & 0.52±0.02 & 0.56±0.01 & 0.523±0.008 & 1.86±0.13 & 33.12±0.15 & 0.852±0.002 & 0.265±0.004 & 40.5±1.1 & 0.39±0.02 & 0.47±0.01 & 0.509±0.007 \\
NVIDIA VSR & 2.57±0.14 & 37.40±0.13 & 0.922±0.002 & 0.082±0.003 & 60.1±1.4 & 0.56±0.02 & 0.60±0.01 & 0.534±0.008 & 2.31±0.12 & 33.55±0.13 & 0.858±0.002 & 0.207±0.003 & 47.8±1.3 & 0.47±0.02 & 0.50±0.01 & 0.518±0.007 \\
XLSR & N/A & N/A & N/A & N/A & N/A & N/A & N/A & N/A & 2.13±0.14 & 33.44±0.14 & 0.855±0.003 & 0.263±0.004 & 43.9±1.4 & 0.42±0.024 & 0.48±0.01 & 0.510±0.007 \\
XLSR$^{T}$ & 2.56±0.12 & 37.25±0.14 & 0.920±0.002 & 0.230±0.010 & 53.4±1.1 & 0.47±0.02 & 0.51±0.01 & 0.515±0.008 & 2.17±0.12 & 33.44±0.12 & 0.856±0.002 & 0.263±0.010 & 44.6±1.2 & 0.43±0.02 & 0.49±0.01 & 0.513±0.007 \\
SMFANet & N/A & N/A & N/A & N/A & N/A & N/A & N/A & N/A & 1.98±0.16 & 33.05±0.16 & 0.850±0.003 & 0.191±0.005 & 43.6±1.4 & 0.42±0.024 & 0.47±0.01 & 0.508±0.007 \\
SMFANet$^{T}$ & 2.37±0.14 & 37.14±0.15 & 0.919±0.002 & 0.158±0.007 & 56.4±1.2 & 0.51±0.02 & 0.55±0.01 & 0.520±0.008 & 2.02±0.14 & 33.06±0.14 & 0.851±0.002 & 0.183±0.007 & 44.3±1.2 & 0.43±0.02 & 0.48±0.01 & 0.511±0.007 \\
SAFMN & N/A & N/A & N/A & N/A & N/A & N/A & N/A & N/A & 1.91±0.17 & 33.36±0.17 & 0.853±0.003 & 0.204±0.005 & 43.4±1.3 & 0.41±0.024 & 0.46±0.01 & 0.505±0.007 \\
SAFMN$^{T}$ & 2.26±0.15 & 37.09±0.15 & 0.918±0.002 & 0.122±0.006 & 57.9±1.3 & 0.53±0.02 & 0.57±0.01 & 0.525±0.008 & 1.95±0.15 & 33.38±0.15 & 0.854±0.002 & 0.199±0.006 & 44.1±1.1 & 0.42±0.02 & 0.47±0.01 & 0.509±0.007 \\
RLFN & 2.17±0.15 & 37.03±0.15 & 0.917±0.002 & 0.086±0.004 & 59.3±1.3 & 0.54±0.02 & 0.58±0.01 & 0.529±0.008 & 2.24±0.16 & \cellcolor{blue!10}{34.42±0.15} & 0.862±0.002 & 0.247±0.004 & 43.9±1.1 & 0.42±0.02 & 0.49±0.01 & 0.515±0.007 \\
RLFN$^{T}$ & 2.69±0.13 & 37.63±0.12 & 0.924±0.002 & 0.072±0.003 & 62.4±1.4 & 0.58±0.02 & 0.62±0.01 & 0.537±0.008 & \cellcolor{blue!10}4.32±0.13 & 33.82±0.12 & 0.860±0.002 & \cellcolor{blue!10}{0.182±0.003} & \cellcolor{blue!10}49.3±1.4 & \cellcolor{blue!10}0.51±0.02 & 0.52±0.01 & 0.522±0.007 \\
SPAN & 2.55±0.12 & 37.45±0.13 & 0.921±0.002 & 0.066±0.003 & 61.8±1.3 & 0.57±0.02 & 0.61±0.01 & 0.535±0.008 & 1.87±0.15 & 32.47±0.12 & 0.845±0.002 & 0.377±0.003 & 46.2±1.2 & 0.46±0.02 & 0.51±0.01 & 0.520±0.007 \\
SPAN$^{T}$ & \cellcolor{blue!10}3.13±0.15 & \cellcolor{blue!10}37.73±0.12 & \cellcolor{blue!10}0.925±0.002 & \cellcolor{blue!10}0.063±0.003 & \cellcolor{blue!10}64.5±1.5 & \cellcolor{blue!10}0.61±0.02 & \cellcolor{blue!10}0.65±0.01 & \cellcolor{blue!10}0.547±0.008 & 3.14±0.12 & 33.51±0.12 & 0.859±0.002 & 0.206±0.003 & 48.1±1.3 & 0.49±0.02 & 0.52±0.01 & 0.523±0.007 \\
\textbf{EfRLFN}$^{T}$ & \cellcolor{blue!20}\textbf{3.33±0.14} & \cellcolor{blue!20}\textbf{37.85±0.11} & \cellcolor{blue!20}\textbf{0.928±0.002} & \cellcolor{blue!20}\textbf{0.059±0.003} & \cellcolor{blue!20}\textbf{67.8±1.4} & \cellcolor{blue!20}\textbf{0.65±0.02} & \cellcolor{blue!20}\textbf{0.74±0.01} & \cellcolor{blue!20}\textbf{0.547±0.008} & \cellcolor{blue!20}\textbf{4.52±0.13} & \cellcolor{blue!20}\textbf{34.55±0.11} & \cellcolor{blue!20}\textbf{0.865±0.002} & \cellcolor{blue!20}\textbf{0.173±0.003} & \cellcolor{blue!20}\textbf{52.7±1.5} & \cellcolor{blue!20}\textbf{0.58±0.02} & \cellcolor{blue!20}\textbf{0.539±0.01} & \cellcolor{blue!20}\textbf{0.527±0.007} \\
\midrule
\multicolumn{17}{l}{\textit{Non-real-time SR models}} \\
Real-ESRGAN & 3.87±0.13 & 37.65±0.12 & 0.926±0.002 & 0.240±0.007 & 68.1±1.2 & 0.66±0.02 & 0.74±0.02 & 0.5485±0.008 & - & - & - & - & - & - & - & - \\
SwinIR & 4.33±0.15 & 37.88±0.11 & 0.928±0.003 & 0.224±0.010 & 69.5±1.4 & 0.67±0.03 & 0.75±0.01 & 0.5505±0.008 & - & - & - & - & - & - & - & - \\
Omni-SR & 4.50±0.14 & 37.92±0.15 & 0.929±0.002 & 0.218±0.006 & 70.2±1.2 & 0.68±0.01 & 0.76±0.01 & 0.5525±0.009 & - & - & - & - & - & - & - & - \\
BasicVSR++ & 4.87±0.13 & 38.05±0.13 & 0.930±0.002 & 0.203±0.007 & 71.8±1.1 & 0.70±0.02 & 0.77±0.01 & 0.5585±0.007 & - & - & - & - & - & - & - & - \\
RVRT & 4.98±0.14 & 38.12±0.13 & 0.931±0.003 & 0.183±0.008 & 72.4±1.2 & 0.71±0.02 & 0.78±0.02 & 0.5625±0.008 & - & - & - & - & - & - & - & - \\
\bottomrule
\end{tabular}
\label{tab:full_results_rotate}
\end{sidewaystable}

\FloatBarrier

\section{Dataset Preview}
\label{app:a12-dataset-preview}

\begin{figure}[h!]
\centering
\includegraphics[width=1.0\linewidth]{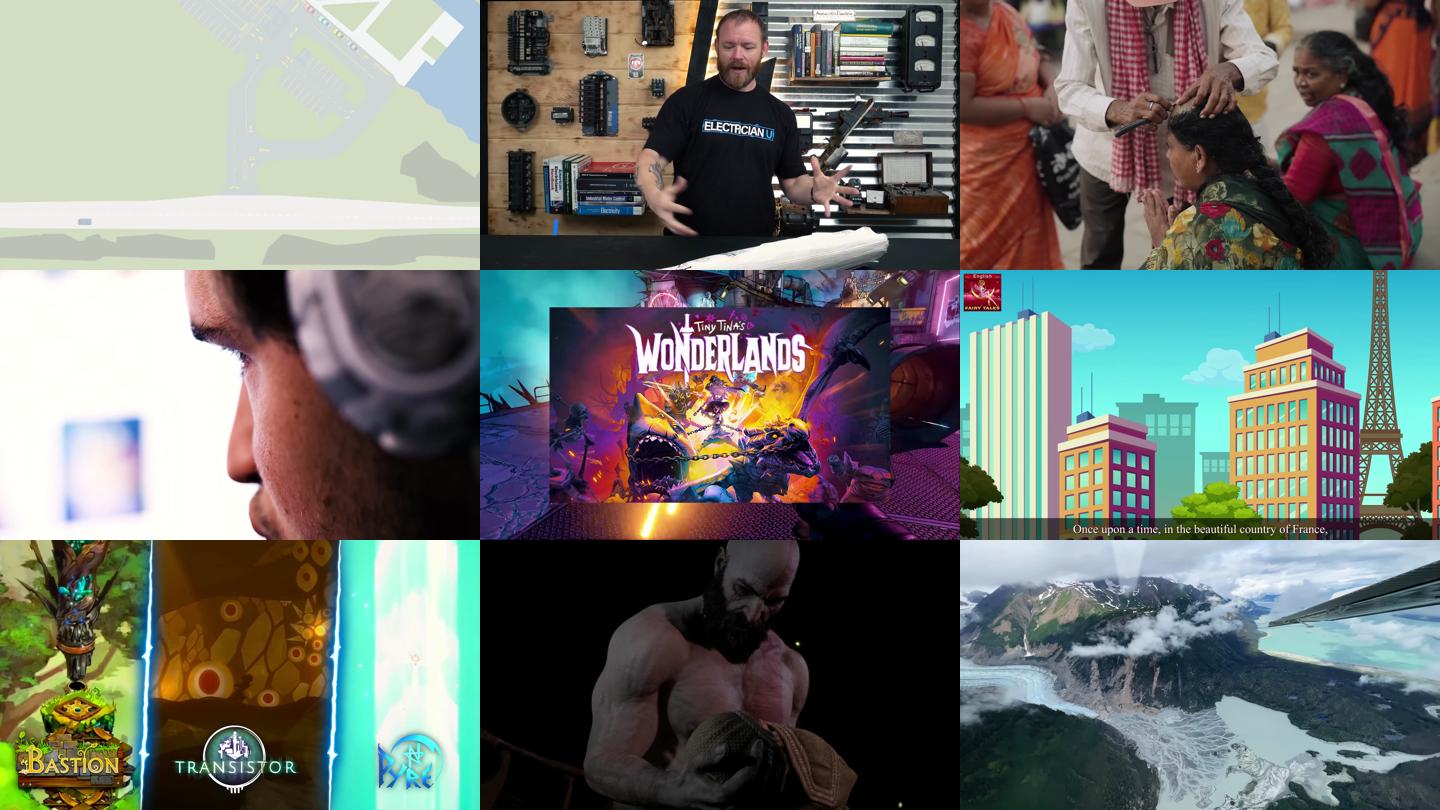}
\caption{Example frames from the proposed StreamSR dataset.}
\label{fig:dataset-frames}
\end{figure}

Figure~\ref{fig:dataset-frames} presents representative frames from the proposed StreamSR dataset. These samples illustrate the diversity of motion patterns, scene types, and compression characteristics captured in the data.

\end{document}